%% file: stylecarigan.tex
\documentclass[acmtog]{acmart}

\usepackage{booktabs}

\usepackage{multirow}
\usepackage{booktabs}
\usepackage{makecell}
\usepackage[export]{adjustbox}
\usepackage{upquote}
\usepackage{stackengine} 
\usepackage{scalerel}
\usepackage{soul}
\usepackage{tabularx}
\usepackage{graphbox}
\usepackage{arydshln}
\usepackage{wrapfig}
\newcolumntype{Y}{>{\centering\arraybackslash}X}

\input{macro}

\usepackage[ruled]{algorithm2e}

\SetAlFnt{\small}
\SetAlCapFnt{\small}
\SetAlCapNameFnt{\small}
\SetAlCapHSkip{0pt}

\setcopyright{acmlicensed}\acmJournal{TOG}
\acmYear{2021}\acmVolume{40}\acmNumber{4}\acmArticle{116}\acmMonth{8} \acmDOI{10.1145/3450626.3459860}

\begin{document}
\title{StyleCariGAN: Caricature Generation via StyleGAN Feature Map Modulation}

\author{Wonjong Jang}
\affiliation{
  \institution{POSTECH}
  \city{Pohang}
  \country{Republic of Korea}}
\email{wonjong@postech.ac.kr}

\author{Gwangjin Ju}
\affiliation{
  \institution{POSTECH}
  \city{Pohang}
  \country{Republic of Korea}}
\email{gwangjin@postech.ac.kr}

\author{Yucheol Jung}
\affiliation{
  \institution{POSTECH}
  \city{Pohang}
  \country{Republic of Korea}}
\email{ycjung@postech.ac.kr}

\author{Jiaolong Yang}
\affiliation{
  \institution{Microsoft Research Asia}
  \city{Beijing}
  \country{China}
}
\email{jiaoyan@microsoft.com}

\author{Xin Tong}
\affiliation{
  \institution{Microsoft Research Asia}
  \city{Beijing}
  \country{China}
}
\email{xtong@microsoft.com}

\author{Seungyong Lee}
\affiliation{
  \institution{POSTECH}
  \city{Pohang}
  \country{Republic of Korea}
}
\email{leesy@postech.ac.kr}

\begin{abstract}
    We present a caricature generation framework based on shape and style manipulation using StyleGAN. Our framework, dubbed {\em StyleCariGAN}, automatically creates a realistic and detailed caricature from an input photo with optional controls on shape exaggeration degree and color stylization type. The key component of our method is shape exaggeration blocks that are used for modulating coarse layer feature maps of StyleGAN to produce desirable caricature shape exaggerations. We first build a layer-mixed StyleGAN for photo-to-caricature style conversion by swapping fine layers of the StyleGAN for photos to the corresponding layers of the StyleGAN trained to generate caricatures. Given an input photo, the layer-mixed model produces detailed color stylization for a caricature but without shape exaggerations. We then append shape exaggeration blocks to the coarse layers of the layer-mixed model and train the blocks to create shape exaggerations while preserving the characteristic appearances of the input. Experimental results show that our StyleCariGAN generates realistic and detailed caricatures compared to the current state-of-the-art methods. We demonstrate StyleCariGAN also supports other StyleGAN-based image manipulations, such as facial expression control.

\end{abstract}

\ccsdesc[300]{Computing methodologies~Image processing}

\keywords{2D caricature, StyleGAN, shape exaggeration block, layer-swapping}

\begin{teaserfigure}
    \centering
    \includegraphics[width=1.0\textwidth]{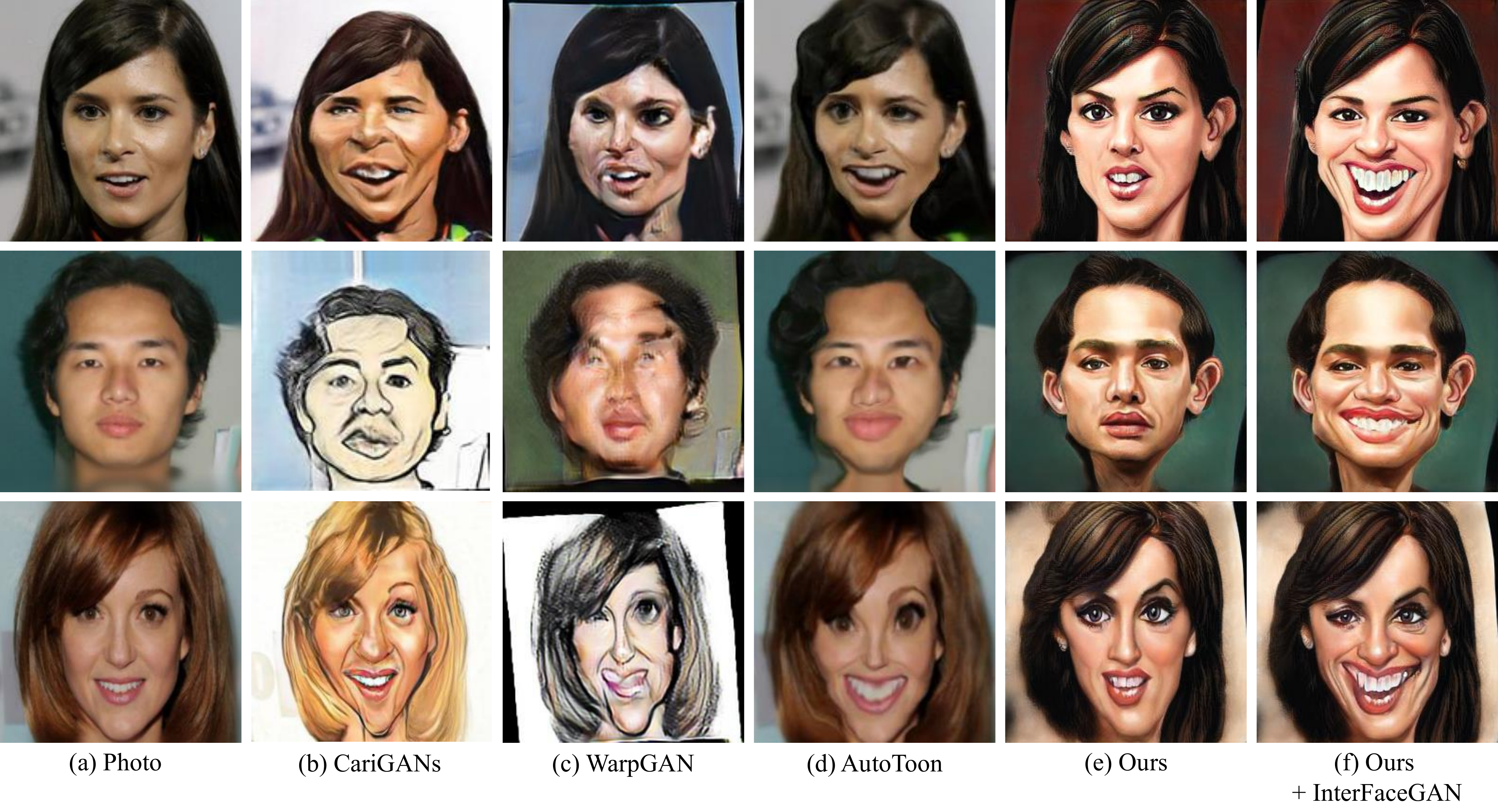}
    \vspace{-0.8cm}
    \label{fig:teaser}
    \caption{\textit{Comparison of previous caricature generation methods with ours.} Different state-of-the-art caricature generation methods are compared to our proposed StyleCariGAN; CariGANs \cite{cao2018carigans} (b); WarpGAN \cite{shi2019warpgan} (c); AutoToon \cite{gong2020autotoon} (d). Our method (e) can be used with other StyleGAN-based image editing methods, e.g., InterFaceGAN \cite{shen2020interfacegan} (f).
    {\footnotesize Photos: \copyright John Roaux/AP Photo, \cite{chen2002pictoon}, \copyright Kevin Winter/Getty Images.}}
\end{teaserfigure}

\maketitle

\input{tex/1.intro}

\input{tex/2.related}

\input{tex/3.method}

\input{tex/4.experiment}

\input{tex/5.conclusion}

\begin{acks}
We would like to thank anonymous reviewers for constructive comments.
This work was supported by the Ministry of Science and ICT, Korea, through
IITP grants (SW Star Lab, IITP-2015-0-00174; Collaborative Research Project with Microsoft Research Asia, IITP-2020-0-01649; Artificial Intelligence Graduate School Program (POSTECH), IITP-2019-0-01906) and MSRA Collaborative Research Grant.
\end{acks}

\bibliographystyle{ACM-Reference-Format}
\bibliography{stylecarigan}

\end{document}

%% file: macro.tex
\newcommand{\Fig}[1] {Fig.~\ref{fig:#1}}
\newcommand{\Figs}[1]{Figs.~\ref{fig:#1}}
\newcommand{\Tbl}[1]  {Table~\ref{tbl:#1}}

\newcommand{\Sec}[1] {Sec.~\ref{sec:#1}}

\newcommand{\Etal}   {{\textit{et al.}}}

\newcommand{\change}[1]{#1}
\newcommand{\blue}[1]{#1}

%% file: tex/1.intro.tex
\section{Introduction}
\label{sec:Intorduction}
A caricature is a type of portrait wherein artists exaggerate the most recognizable characteristics of their subjects while oversimplifying other characteristics.
To draw a caricature, artists have to learn how to not only paint in cartoon style but also capture and exaggerate the most salient facial features, which requires long time of training for development.
Even after developing the skills, artists can take hours or even days to draw a single piece of caricature.
Automatic computational strategies for photo-to-caricature translation can reduce the training and production burdens and make caricatures readily available for the general public.

Early studies in caricature generation \cite{akleman1997making, akleman2000making, gooch2004human} suggested methods to capture and deform facial geometric features, but they relied on user interaction to produce impressive results. Several automatic caricature generation methods were proposed, but their artistic styles were limited to pre-defined rules \cite{brennan1985caricature, le2011shape, mo2004improved}. 

Recent deep learning algorithms for image-to-image translation can find a mapping from an input domain to an output domain given training examples, but giving a proper supervision to learn automatic photo-to-caricature translation is not trivial. A straightforward way would be to collect photo-caricature pairs,
but the challenge is that photos and caricatures only have weak correspondences. For example, the caricatures corresponding to one person can vary in pose, expression, and exaggeration style. Constructing pairs using only the identity matching would make a sparse dataset for training photo-to-caricature translation. 

Unpaired image-to-image translation algorithms such as CycleGAN \cite{zhu2017unpaired} can be trained without explicit pairing between input and output training examples. However, these unpaired image-to-image translation approaches still have a challenge when crossing domains using a sparse dataset such as photo and caricature dataset. 
Simply using a generic unpaired image-to-image translation approach does not result in realistic and visually pleasing results, as demonstrated in \cite{cao2018carigans}.
It seems the supervision between the photo domain and the caricature domain must be specifically crafted because of large shape variations between the two domains.

Recent automatic caricature algorithms design specialized network architectures for photo-to-caricature translation by separating geometric deformation and texture stylization \cite{cao2018carigans, shi2019warpgan, gong2020autotoon}. However, the geometric deformation modules are based on 2D image warping and usually rely on interpolation of sparse control points \cite{cao2018carigans, shi2019warpgan}, resulting in loss of detailed shape deformations present in real caricatures. Collecting a small amount of dense and detailed caricature shape deformation examples was studied \cite{gong2020autotoon}, but collecting a large amount of such dense data would be laborious.

To create realistic and detailed deformations, we take a generative approach that produces shape variations at multiple scales by modulating the corresponding feature maps in the network. By moving away from 2D image warping, we can handle a wide range of deformations from changes of overall facial shape to addition of wrinkles.
Specifically, in this paper, we use StyleGAN \cite{karras2019style} to manipulate a rich space of shape deformations in an unsupervised way. 
StyleGAN generates intermediate feature maps of multiple scales, where each scale represents a different spatial scale of the modeled image. It was showcased that StyleGAN can separately control over coarse, medium, and fine scale facial attributes.

One challenge for using StyleGAN as a photo-to-caricature translator is that it models the distribution of only one domain.
We can train two StyleGANs separately for facial photos and caricature images. 
Then, as our translator must take a photo as the starting point, we need a way to connect manipulated photo feature maps to caricature styles in the final image.
We resolve the problem 
by simply swapping layers of two StyleGANs,
as in \cite{pinkney2020blending}. Our layer-mixed StyleGAN uses the first four layers of the photo StyleGAN and the last three layers of the caricature StyleGAN. 
Another challenge for using StyleGAN to build a system that takes a photo as the input is the encoding of the input photo into a StyleGAN latent vector. We combine two previous ideas on GAN inversion \cite{tewari2020pie, karras2020analyzing} to obtain a latent code optimized for the input photo.

Given the latent code of an input photo, our layer-mixed StyleGAN can create an output image with caricature color styles. However, the output would not contain enough shape exaggerations as the facial shape is dominated by the first four layers copied from the photo StyleGAN.
We then append \textit{shape exaggeration blocks} to the four layers of the layer-mixed StyleGAN to handle shape deformations with feature map modulation.
Our shape exaggeration blocks compute desirable feature map changes to be added to the original feature maps, and provides multi-scale control for shape exaggerations.

There are desirable properties for shape exaggeration blocks: 1) The resulting deformations should generate an image that looks like a realistic caricature. 2) The deformations should preserve unique visual features of the input photo. The first property can be achieved by adversarial training using a caricature dataset. However, the second property is not trivial to achieve. To design a constraint on preservation of important features, we need a tool to extract a set of visual features shared between photos and caricatures. The extraction is challenging because of large shape and texture gaps between the two domains.

Previous photo-to-caricature translators based on deep learning solved the problem of preservation of visual features in different ways. Cao~\Etal~\shortcite{cao2018carigans} designed a \textit{characteristic loss} to constrain their 2D image warping using the cosine similarity between input and output landmark displacements,
and the loss was used along with a cycle consistency loss between a photo and a caricature. 
Shi \Etal~\shortcite{shi2019warpgan} designed an \textit{identity-preserving adversarial loss} that makes the discriminator for the adversarial training also perform face identification between a photo and a caricature.

\begin{table*}[t]
\centering
\normalsize
\caption{\textit{Comparison of state-of-the-art deep learning based 2D automatic caricature generation methods and ours.}}
\vspace*{-0.1cm}
\begin{tabularx}{2\columnwidth}{c@{\hspace{0pt}}c Y Y Y Y}
\toprule
& & CariGANs & WarpGAN & AutoToon & Ours\\
&  & \cite{cao2018carigans} & \cite{shi2019warpgan} & \cite{gong2020autotoon} & 
\\
\midrule
Shape manipulation & & 2D spatial interpolation & 2D spatial interpolation & 2D spatial dense & StyleGAN 
\\
type &  & of landmark & of control point & displacement map & feature map modulation 
\\
&  & deformations    & deformations  &  &  
\\
\\
Shape manipulation & & Pre-defined & Dynamically decided & Dense & Dense
\\ 
spatial density & & 63 landmarks detected &              16 control points &  & 
\\ 
 & & from input image &              & & 
\\
\\
Training & & Unpaired large & Unpaired large & Paired small & Unpaired large
\\
dataset type& & (10K photos + & (6K photos + & (101 photos +  & (70K photos +
\\
 & & <10K caricatures) & <10K caricatures) & 101 caricatures) & <10K caricatures)
\\
\\
User control & & single exaggeration & single exaggeration & single exaggeration & 4-scale exaggeration
\\
 & & magnitude control & magnitude control & magnitude control & magnitude control
\\
\bottomrule

\end{tabularx}

\label{tbl:summary}
\end{table*}

Our approach to enable the shape exaggeration blocks to preserve the visual features of an input image is based on two key ideas. 1) We establish weak correspondence between photos and caricatures in an unsupervised way with {\em cycle consistency}, as in \cite{zhu2017unpaired} and \cite{cao2018carigans}. For the cycle consistency, we create two models for photo-to-caricature translation and caricature-to-photo translation. 2) We guide the shape exaggeration blocks with a direct supervision for preservation of facial attributes using a recently released dataset, WebCariA \cite{HuoECCV2020WebCariA}, which provides a shared set of annotations between photos and caricatures. WebCariA provides 50 label annotations based on intrinsic facial attributes. The annotations include shape attribute labels of overall facial shapes, nose shapes, eye shapes, and mouth shapes, i.e., \textit{Wide Nose, Mustache, Big Eyes, Arched Eyebrows, etc}, and are provided both for photo and caricature images. Using the dataset, we train two attribute classifiers for photos and caricatures separately. We then design an {\em attribute matching loss} that constrains the attributes detected in the input photo to be preserved in the output caricature.

To summarize, our main contributions are as follows:
\begin{itemize}
    \item{We propose a novel {\em StyleCariGAN} framework for photo-to-caricature translation. Based on StyleGAN, it manipulates feature maps at multiple scales to produce desired shape exaggeration and color stylization.}
    \item{To create desirable shape exaggerations in photo-to-caricature translation, we propose shape exaggeration blocks that are appended to the coarse layers of a layer-mixed StyleGAN.}
    \item{We design an attribute matching loss that is used along with cycle consistency to constrain the shape exaggeration blocks to preserve important visual features of the input.}
    \item{Our StyleCariGAN framework generates realistic and detailed caricatures compared to state-of-the-art methods, supporting other image manipulations such as facial expression control.}
\end{itemize}

%% file: tex/2.related.tex
\section{Related Work}

\subsection{Rule-based automatic caricature generation}

Rule-based caricature generation methods create caricatures by transforming an input image with a fixed sequence of operations. The seminal work of Brennan~\shortcite{brennan1985caricature} suggested an automatic algorithm based on exaggeration of relative feature point differences between an input face and a template mean face. This idea was further explored in many other algorithms. Mo~\Etal~\shortcite{mo2004improved} extended it by considering the variances of different feature point positions. Liao~\Etal~\shortcite{liao2004automatic} exaggerated facial feature points constructed in a tree structure. Le~\Etal~\shortcite{le2011shape} warped an input face based on exaggeration of anthropometric ratios between facial components. 

Although these rule-based methods create plausible caricature deformations based on carefully designed procedures, the procedurally deformed facial shape may not resemble a realistic and artist-drawn caricature.
The designed procedures cannot cover a wide range of caricature deformations performed by human artists. 
A natural solution to this problem would be modeling the mapping from a photo to a caricature in an end-to-end fashion, as done in recent deep learning based image translation methods.

\subsection{Deep image-to-image translation}

With the advance of deep convolutional neural networks, image-to-image translation methods have been developed to convert an image in the source domain to the target domain.
Given paired images from the two domains as training data, Isola~\Etal~\shortcite{isola2017image} and Wang~\Etal~\shortcite{wang2018high} showcase end-to-end training schemes for image translation based on adversarial training~\cite{goodfellow2014generative}. These algorithms work well when input-output pairs are well-defined. Using unpaired data for image-to-image translation learning has also been actively studied. CycleGAN~\cite{zhu2017unpaired} learns the mapping between two image domains without paired correspondence using a cycle consistency constraint. U-GAT-IT \cite{kim2019u} learns the mapping using attention and a new feature normalization method. MUNIT~\cite{huang2018multimodal} considers the fact that there can be multiple outputs possible for a given input and proposes a multimodal translation framework.

The difficulty of applying image-to-image translation to caricature generation comes from lack of abundant identity-matched photo-caricature pairs. Moreover, the two domains of real photos and caricatures differ at not only image style, but also face shapes. Naively applying unpaired translation would result in inferior caricature quality and less identity preservation, as shown in \cite{cao2018carigans} . 
In this paper, we do not explicitly pair photos and caricatures, but supervise the correspondence between the two with constraint on facial attribute preservation. In addition, we make the mapping be learned more easily with a decomposition in caricature image formation: shape manipulation and appearance stylization.

\vspace{-0.13cm}
\subsection{Deep caricature generation}

Automatic caricature generation algorithms based on deep learning has been studied recently.
Cao~\Etal~\shortcite{cao2018carigans} proposed CariGANs for photo-to-caricature translation using two networks for geometric deformation and style transfer.
They constrain the output caricature to preserve visual features of the input shape by applying a cycle consistency and a cosine similarity constraint to landmark deformations. For more flexible shape variation, WarpGAN \cite{shi2019warpgan} trained an image warping module with dynamically decided control points using deep neural networks. WarpGAN created caricatures that resemble the input by making the discriminator used for the adversarial learning perform identity classification. AutoToon~\cite{gong2020autotoon} adopted dense deformation fields, instead of using coarse control points, and trained the caricature generator using a small amount of paired dataset comprising photos and corresponding caricatures generated by artists using 2D deformation tools. 
\change{CariGAN \cite{li2020carigan} proposed an image fusion mechanism to encourage the caricature generator to focus on important local regions around sparse facial landmarks.}
3D caricature algorithms can be used to create a 2D caricature by warping the input image based on the created 3D shape. Han~\Etal~\shortcite{han2018caricatureshop} create 3D caricatures with deep neural networks by generating and manipulating a rough sketch of an exaggerated 3D facial shape. The final 2D caricature image is generated by blending a warped image based on the 3D warping and the original image. 

Our main difference to previous deep-learning-based caricature generation methods is that we take a generative approach for caricature exaggerations by modeling spatially dense and detailed shape deformations using a large amount of unpaired photo-caricature data. We create realistic and detailed caricatures by obtaining desirable shape deformations through modulations of StyleGAN feature maps. We also provide a multi-scale control over shape exaggeration magnitude in the caricature output.
\Tbl{summary} summarizes the differences among caricature generation methods.

\subsection{Image generation and editing using StyleGAN}

StyleGAN \cite{karras2019style, karras2020analyzing} is a powerful generative image model, which can synthesize high-resolution face images that are even hard to distinguish from real photos. It is shown that the feature maps at different resolutions of the StyleGAN architecture characterize the image styles at different scales and control the generation in a coarse-to-fine fashion. With StyleGAN, image manipulation can be achieved by embedding a real photo into the StyleGAN latent space and editing the embedded latent code for re-generation \cite{abdalimage2stylegan, abdal2020image2stylegan++, zhu2020domain, deng2020disentangled, tewari2020pie, karras2020analyzing}. 
\change{Semantically-meaningful editing in GAN latent space has also been studied \cite{zhu2016generative, brock2016neural, yeh2017semantic, creswell2018inverting, bau2019semantic, richardson2020encoding}. Recent studies based on StyleGAN achieve such editing via either supervised  \cite{shen2020interfacegan, abdal2020styleflow} or unsupervised  \cite{shen2020closed} learning.}

\textit{Toonification} \cite{pinkney2020blending} creates an image with cartoon structure and photo-realistic rendering using StyleGAN. It swaps layers of two StyleGANs trained for photos and caricatures. The StyleGAN for photos is trained from scratch and the StyleGAN for caricatures is fine-tuned from the photo model. It uses the fine-tuned caricature StyleGAN for coarse layers to create cartoon structures, and the photo StyleGAN for fine layers to create photo-realistic textures. 
To generate a \textit{toonified} image from an input photo, separately generated cartoon structures are blended into the input photo feature maps. \change{FreezeG \cite{freezeg2020github} achieves similar pseudo image translation through training instead of layer-swapping. It freezes a certain range of layers in a StyleGAN for one domain, and fine-tunes the StyleGAN to generate images of another domain. However, in both Toonification and FreezeG, it is unclear how to explicitly supervise the translations to preserve visual features of input images.}

Our automatic caricature system generates an image with exaggerated photo structure and cartoon-style rendering. The exaggerated photo structure is created from the input, preserving important visual features. We handle the problem of crossing the domains from photo to caricature by swapping layers of two StyleGANs, similarly to Toonification \cite{pinkney2020blending}. However, differently from Toonification, the coarse layers are copied from the photo StyleGAN and the fine layers are from the caricature StyleGAN in our method. Moreover, we append learnable shape exaggeration blocks to the copied coarse layers and design a supervision for the blocks to preserve characteristic attributes of the input. Comparison between our method and Toonification is presented in \Sec{toonification}.

%% file: tex/3.method.tex
\section{Photo-to-caricature translation framework}

\begin{figure}[t]
    \centering
    \includegraphics[width=0.4\textwidth]{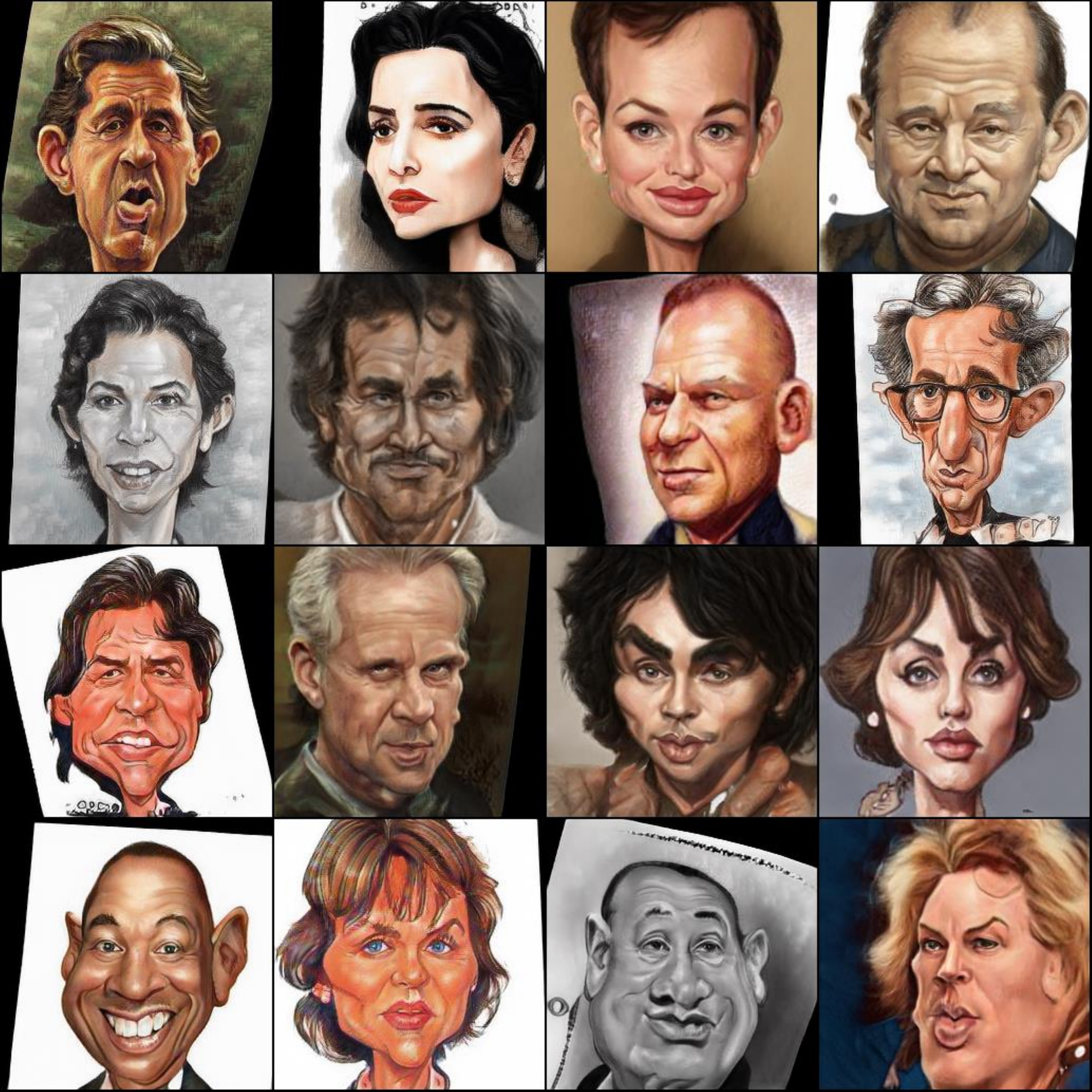}
    \vspace*{-0.15cm}
    \caption{
    \textit{Image samples generated using the fine-tuned StyleGAN for caricatures.\protect\footnotemark} To sample the images, we apply the truncation trick \cite{karras2019style, karras2020analyzing} with $\psi = 0.7$.} 
    \vspace*{-0.3cm}
    \label{fig:generated}
\end{figure}

Our StyleCariGAN framework contains two fixed StyleGANs: one is trained to generate photos, and the other is fine-tuned from the photo model to generate caricature images (\Fig{generated}). After pre-training, the two models are fixed and not updated during our process. We refer to the StyleGAN trained to generate regular face images as the \textit{photo StyleGAN} and another trained for caricature images as the \textit{caricature StyleGAN}. 
By swapping layers of the two StyleGANs, we make a layer-mixed model that generates a caricature image starting 
from a latent code.

In the layer-mixed StyleGAN model, the coarse (low-resolution) layers are from photo StyleGAN to control the overall structural and identity information of the output caricature. The fine (high-resolution) layers are from caricature StyleGAN to create detailed color styles for a caricature.
However, naively copying the coarse layers from photo StyleGAN would produce ordinary face shapes without exaggerations.

To impose shape exaggeration capability on the layer-mixed model, we append \textit{shape exaggeration blocks} to the coarse layers. 
Our key idea is to train the shape exaggeration blocks with a specially crafted supervision for photo-to-caricature translation. 
We train the blocks to introduce feature map modulations that create realistic and identity-preserving caricature exaggerations. \change{Shape exaggeration blocks implicitly increase the deviation from the average facial geometry in feature space because our model is trained to generate highly diverse caricature geometry with GAN loss. We design cycle consistency and attribute matching losses to guide the deviation to be \blue{distinctively} reflecting the features of the input.}

The starting point of photo-to-caricature translation, {\textit{i.e.}}, the input to the layer-mixed model, is obtained from the input photo using photo-to-latent embedding.
We obtain the $\mathcal{W}+$ latent code \cite{abdalimage2stylegan} representing the input image through iterative optimization, akin to \cite{tewari2020pie} and \cite{karras2020analyzing}.
From the $\mathcal{W}+$ code, our layer-mixed model generates multi-scale feature maps and eventually the output caricature image.

\footnotetext{\change{Since we added zero-padding on borders of training images after aligning the caricature dataset with facial landmarks, black borders sometimes appear in the results.}}

\begin{figure}[t]
    \centering
    \includegraphics[width=0.45\textwidth]{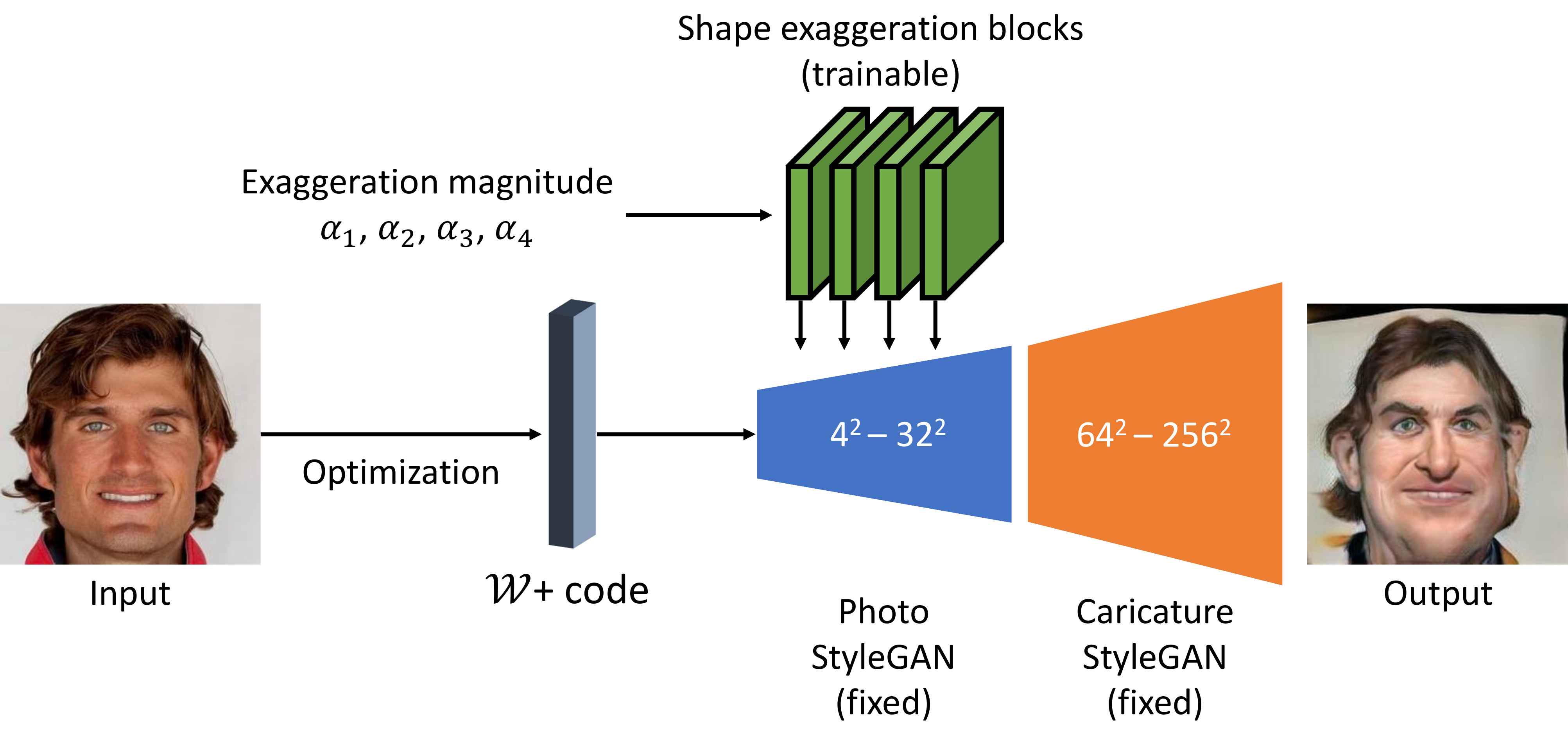}
    \vspace{-8pt}
    \caption{\textit{Overview of our approach for caricature generation using a layer-mixed StyleGAN with shape exaggeration blocks.} A $\mathcal{W}+$ code is first obtained by embedding an input to the latent space of photo StyleGAN. Then, the first four coarse feature maps generated by the layers copied from photo StyleGAN using the $\mathcal{W}+$ code are modulated with shape exaggeration blocks. Finally, the modulated features are fed to the fine layers copied from caricature StyleGAN to produce the output caricature.
    }
    \vspace*{-0.5cm}
    \label{fig:overview}
\end{figure}

\Fig{overview} shows the overall process of our photo-to-caricature translation. The following subsections explain each step of the process in detail.

\begin{figure*}[t]
    \centering
    \small
    \begin{tabularx}{\textwidth}{c@{\hspace{7pt}}c@{\hspace{0pt}}c@{\hspace{0pt}}c@{\hspace{0pt}}c@{\hspace{0pt}}c@{\hspace{0pt}}c@{\hspace{7pt}}c}
    \vspace{0pt}
    \includegraphics[width=0.12\textwidth]{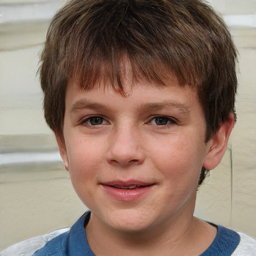} &
    \includegraphics[width=0.12\textwidth]{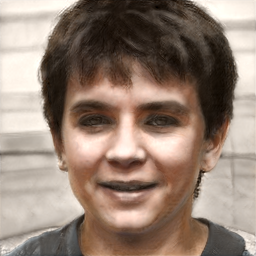} &
    \includegraphics[width=0.12\textwidth]{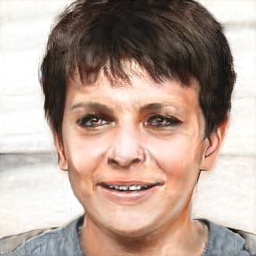} &
    \includegraphics[width=0.12\textwidth]{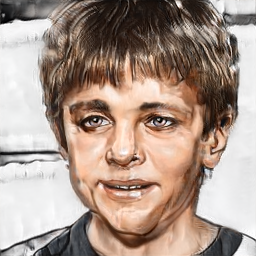} &
    \includegraphics[width=0.12\textwidth]{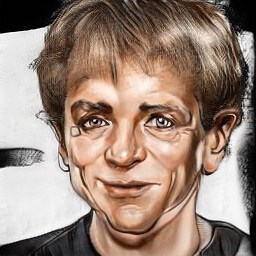} &
    \includegraphics[width=0.12\textwidth]{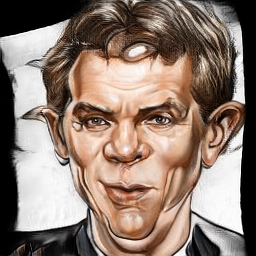} &
    \includegraphics[width=0.12\textwidth]{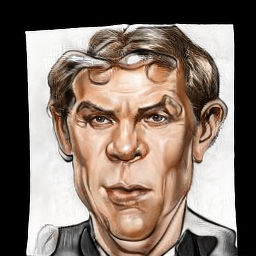} &
    \includegraphics[width=0.12\textwidth]{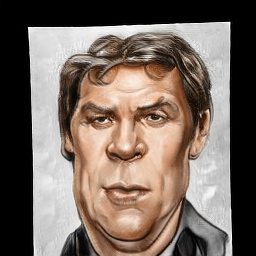} \\
    \includegraphics[width=0.12\textwidth]{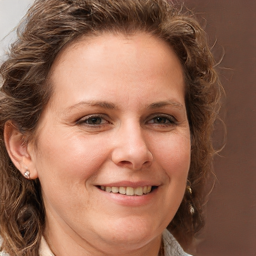} &
    \includegraphics[width=0.12\textwidth]{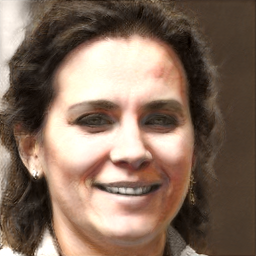} &
    \includegraphics[width=0.12\textwidth]{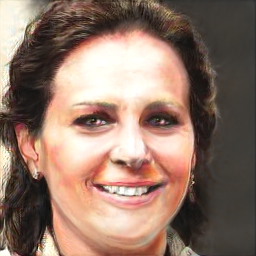} &
    \includegraphics[width=0.12\textwidth]{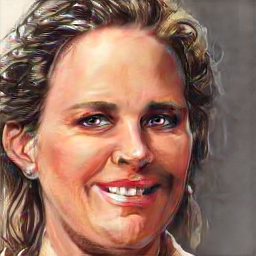} &
    \includegraphics[width=0.12\textwidth]{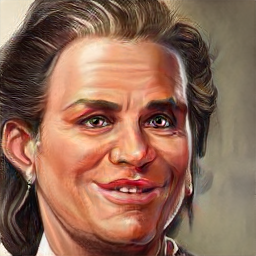} &
    \includegraphics[width=0.12\textwidth]{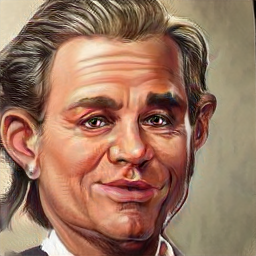} &
    \includegraphics[width=0.12\textwidth]{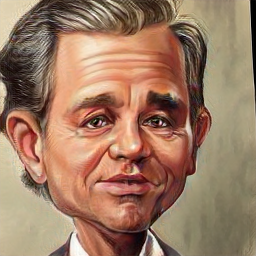} &
    \includegraphics[width=0.12\textwidth]{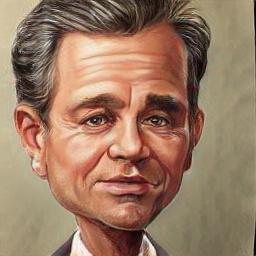} \\
    (a) p = 7, c = 0 & (b) p = 6, c = 1 & (c) p = 5, c = 2 & (d) \underline{\textbf{p = 4, c = 3}} & (e) p = 3, c = 4 & (f) p = 2, c = 5 & (g) p = 1, c = 6 & (h) p = 0, c = 7
    \end{tabularx}
    \vspace{-0.3cm}
    \caption{\textit{Layer swapping examples.} 
    $p$ denotes the number of coarse photo-StyleGAN layers used in the layer-mixed model, and $c$ denotes the number of fine caricature-StyleGAN layers. For instance, $p = 4, c = 3$ means that the first four coarse layers are from the photo StyleGAN and the last three fine layers are from the caricature StyleGAN. As $p$ decreases, the face structure of the input is deformed and the identity is lost. We choose $p = 4, c = 3$ for our layer-mixed model that retains the shape information in the input while producing enough stylization of details.}
    \label{fig:layermix}
    \vspace{-0.1cm}
\end{figure*}

\subsection{Layer swapping between two StyleGANs}
\label{sec:method-layer-mixing}

For the layer-mixed model, we copy the parameters of coarse and fine layers from photo and caricature StyleGANs, respectively. 
\blue{When training caricature StyleGAN with fine-tuning starting from photo StyleGAN, we align caricatures and photos based on facial landmarks. The positions of important visual features are shared between photos and caricatures after the alignment. This spatial alignment enables layer swapping to produce a plausible mixture of a photo and a caricature (\Fig{layermix}).}

In layer swapping, the choice of the boundary between coarse and fine layers is important.
Different choices will lead to different levels of input structure preservation and detail stylization, as shown in \Fig{layermix}. When selecting the boundary, our goal is to enable the layer-mixed model to generate caricature images faithful to the input photo in terms of face structure and identity. We empirically choose the first four scales ($4^{2}\sim32^{2}$) as the coarse layers and the rest as the fine layers.
\Fig{layermix} shows that taking more than three fine layers from caricature StyleGAN would introduce excessive shape deformations that may change \blue{the visual appearance} of the input photo.

The fine layers of our StyleCariGAN take style latent codes that can be selected by a user to control detail styles. For convenient control, we construct a curated set of style latent codes by selecting a number of good examples from randomly generated styles.

The copied parameters of the layers are not updated further after layer swapping. That is, we do not apply additional end-to-end fine-tuning to the layer-mixed model, as the coarse and fine layers of the model already have the desired properties of handling facial shapes and detailed styles, respectively. The only trained components in the layer-mixed model are shape exaggeration blocks appended to the coarse layers. 

To train shape exaggeration blocks, we define another layer-mixed model for caricature-to-photo translation to enable cycled training. Specifically, we construct this new layer-mixed model by taking the first four coarse layers from caricature StyleGAN and the remaining fine layers from photo StyleGAN. We denote this new layer-mixed model as the 
\textit{c2p} model, while the original one, which is the backbone for our StyleCariGAN, is denoted as the \textit{p2c} model.

\subsection{Shape exaggeration blocks}
\label{sec:method-shape-blocks}

Our key component for photo-to-caricature translation is the shape exaggeration blocks.
Four shape exaggeration blocks are attached to the p2c model, and we call the modified p2c model \textit{p2c-StyleCariGAN}, or simply \textit{StyleCariGAN}. We create \textit{c2p-StyleCariGAN} similarly, then use the model for enforcing cycle consistency.

Our shape exaggeration blocks are convolutional layers that produce additive feature modulation maps for the coarse layers of StyleCariGAN. A shape exaggeration block takes a feature map of size $n^2$ and creates a $n^2$ feature modulation map which is added back to the input feature map, akin to residual learning \cite{he2016deep}. The modulated feature map is then fed to the next layer that creates a higher resolution feature map to handle finer-scale structures. The architecture of a shape exaggeration block of size $n^2$ is the same as that of a convolution layer of size $n^2$ in the original StyleGAN \cite{karras2019style}, except that we exclude the upsampling block.

\begin{figure*}[t]
    \centering
    \includegraphics[width=0.78\textwidth]{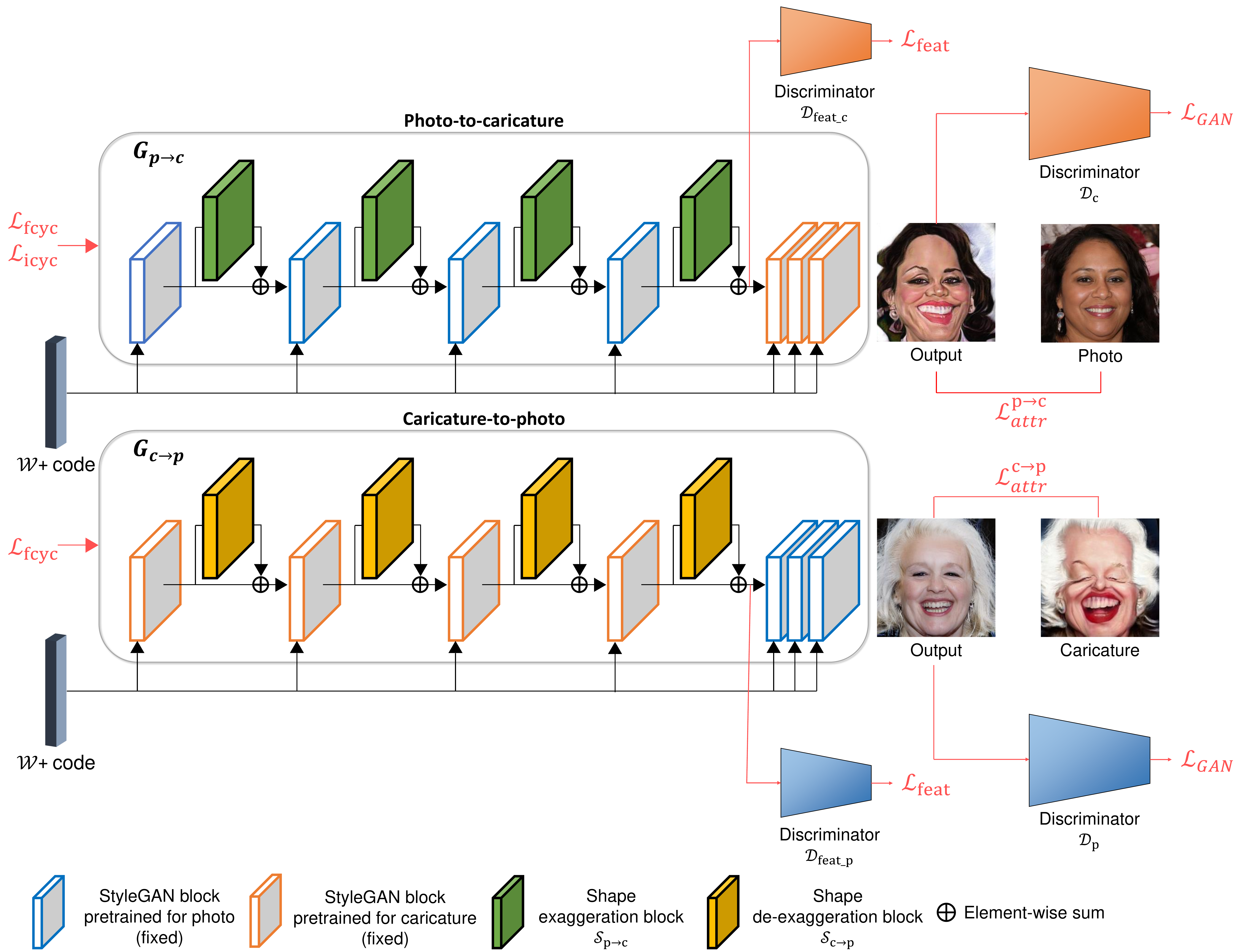}
    \vspace{-2mm}
    \caption{\change{\textit{Training shape exaggeration blocks.} With fixed layers copied from photo StyleGAN and caricature StyleGAN, we train two sets of shape exaggeration blocks: $\delta_{p \rightarrow c}$ in p2c-StyleCariGAN and $\delta_{c \rightarrow p}$ in c2p-StyleCariGAN. 
    To achieve realistic and facial-attribute-preserving exaggerations, shape exaggeration blocks are supervised with the adversarial losses ($\mathcal{L}_{feat}$, $\mathcal{L}_{GAN})$, the cycle consistency losses ($\mathcal{L}_{fcyc}$, $\mathcal{L}_{icyc}$), and the attribute matching loss ($\mathcal{L}_{attr}$).}}
    \label{fig:training}
    \vspace{-1mm}
\end{figure*}

We train the shape exaggeration blocks to achieve two goals. The first goal is to introduce shape deformations that resemble real caricatures, which is achieved by two types of generative adversarial losses: $\mathcal{L}_{feat}$ and $\mathcal{L}_{GAN}$. The second goal is to preserve important visual features from the input photo, which is achieved by three losses: feature map cycle consistency loss $\mathcal{L}_{fcyc}$, identity cycle consistency loss $\mathcal{L}_{icyc}$, and attribute matching loss $\mathcal{L}_{attr}$ (\Fig{training}).

\paragraph{Generative Adversarial Loss}
We use the same generative adversarial loss and supervision method as StyleGAN2 \cite{karras2020analyzing}. For the \textit{real} data needed for the supervision of discriminators, we use generated examples from the caricature StyleGAN, instead of real caricatures. This strategy helps the training process by providing a diverse set of samples and avoiding image loading overhead. One consequence is that our StyleCariGAN would produce caricatures similar to caricature StyleGAN. However, the caricature StyleGAN can create high-quality caricatures as shown in \Fig{generated}, supporting the strategy.
We apply adversarial learning to both the modulated feature maps and final images by introducing two discriminators. We use the non-saturating logistic loss \cite{goodfellow2014generative} with $R_{1}$ regularization \cite{mescheder2018training} for both, and calculate the sum of the two losses as our total adversarial loss $\mathcal{L}_{adv}$:
\begin{equation}
    \mathcal{L}_{adv} = \mathcal{L}_{feat} + \lambda_{GAN}\mathcal{L}_{GAN},
\end{equation}
where $\mathcal{L}_{feat}$ and $\mathcal{L}_{GAN}$ are the losses on feature maps and final images, respectively.

\paragraph{Cycle consistency Loss}

To guide the shape exaggeration blocks to build a correspondence between photos and caricatures, we constrain the blocks to be cycle-consistent. To implement cycle consistency, we utilize the symmetric set of StyleCariGANs: p2c-StyleCariGAN and c2p-StyleCariGAN. We impose cycle consistency both on intermediate modulated feature maps and generated images. We refer to the cycle consistencies for the modulated feature maps and images as feature map cycle consistency $\mathcal{L}_{fcyc}$ and identity cycle consistency $\mathcal{L}_{icyc}$, respectively.

\begin{figure}[h]
    \centering
    \includegraphics[width=0.44\columnwidth]{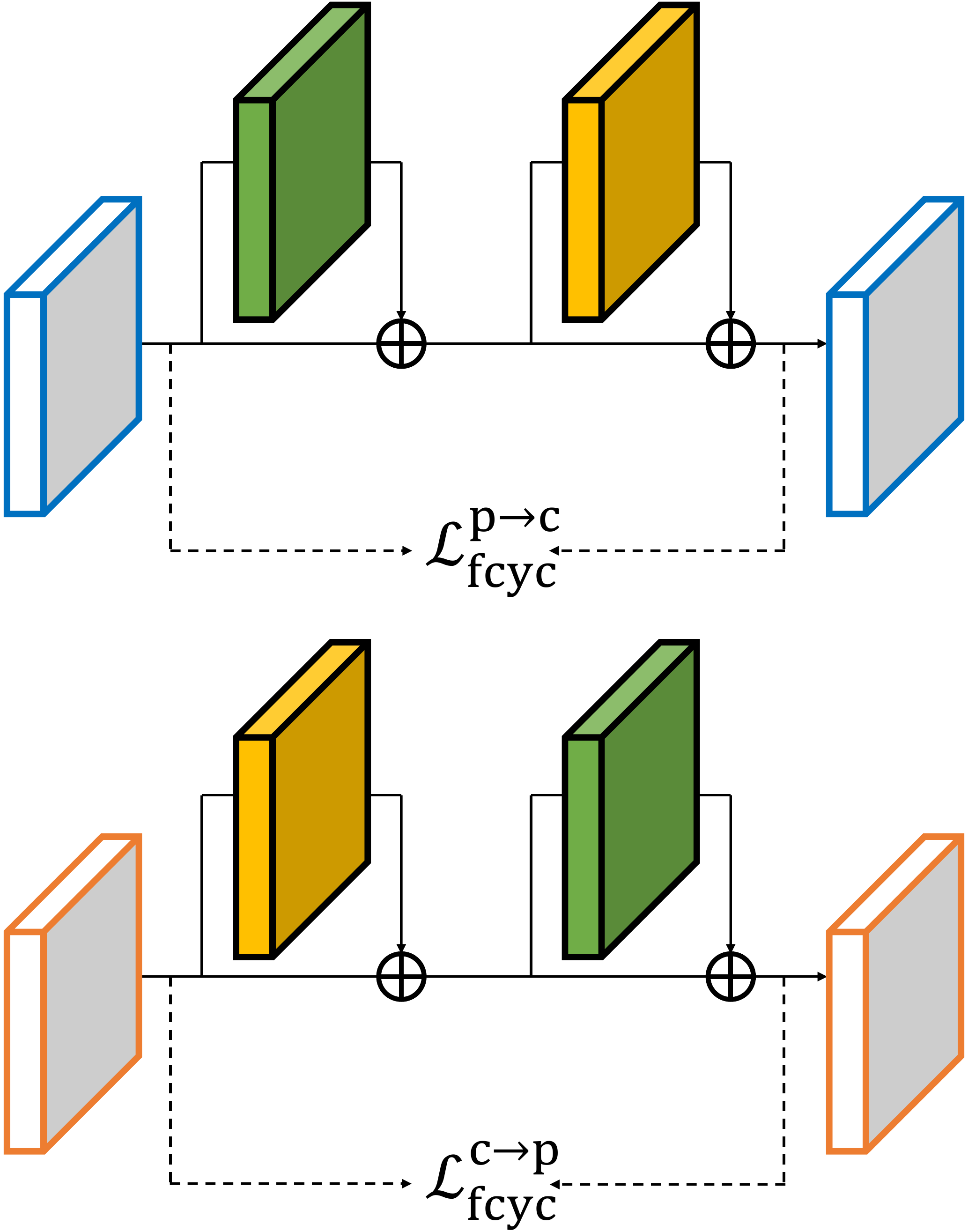}
    \caption{\textit{Features map cycle consistency.} We enforce cycle consistency between the feature maps of coarse layers in p2c-StyleCariGAN and c2p-StyleCariGAN.}
    \vspace*{-0.3cm}
    \label{fig:fcycle}
\end{figure}

The feature map consistency forces the effect of shape exaggeration blocks to be invertible with a cycle at each feature map scale. The shape exaggeration blocks for p2c-StyleCariGAN define photo-to-caricature feature modulation $\mathcal{S}_{p\rightarrow c}$. The corresponding blocks in c2p-StyleCariGAN define caricature-to-photo feature modulation $\mathcal{S}_{c\rightarrow p}$. That is, the blocks in c2p-StyleCariGAN does inverse of exaggeration, reverting an exaggerated shape into a regular one. We call these blocks {\em shape de-exaggeration blocks}. Using the two feature modulations, the cycle consistency loss $\mathcal{L}_{fcyc}$ is defined as:

\begin{gather}
        \mathcal{L}_{fcyc}^{p \rightarrow c} = \sum_{i=1}^{4} ( \mathbb{E}_{\mathcal{F}_{p}^{i}\sim G_{p}^{i}(w)}[\Vert \mathcal{S}^{i}_{c \rightarrow p}(\mathcal{S}^{i}_{p \rightarrow c}(\mathcal{F}_{p}^{i})) - \mathcal{F}_{p}^{i} \Vert_{2}] ),\nonumber\\
        \mathcal{L}_{fcyc}^{c \rightarrow p} = \sum_{i=1}^{4} ( \mathbb{E}_{\mathcal{F}_{c}^{i}\sim G_{c}^{i}(w)}[\Vert \mathcal{S}^{i}_{p \rightarrow c}(\mathcal{S}^{i}_{c \rightarrow p}(\mathcal{F}_{c}^{i})) - \mathcal{F}_{c}^{i} \Vert_{2}] ),\nonumber\\
        \mathcal{L}_{fcyc} = \mathcal{L}_{fcyc}^{p \rightarrow c} + \mathcal{L}_{fcyc}^{c \rightarrow p}, 
\end{gather}
where $F_{c}^{i}$ is a caricature feature map generated by the $i$-th block $G_{c}^{i}$ of the caricature StyleGAN, $F_{p}^{i}$ is a photo feature map of the $i$-th block $G_{p}^{i}$ of the photo StyleGAN, $S_{p\rightarrow c}^{i}$ is the $i$-th photo-to-caricature shape exaggeration block, and $S_{c\rightarrow p}^{i}$ is the $i$-th caricature-to-photo shape de-exaggeration block. \Fig{fcycle} illustrates the losses.

The identity consistency forces the effect of shape exaggeration blocks to be invertible with a cycle by inspecting photos generated from feature maps. 
We design the identity consistency loss $\mathcal{L}_{icyc}$ based on a face embedding network trained on photos \cite{schroff2015facenet}, so the identity consistency is calculated only for the cycle

starting from photo feature maps. Using the face embedding network, 
we design the identity cycle consistency loss $L_{icyc}$ as the $L_2$ distance between two embeddings: 1) the face embedding of a starting photo and 2) the face embedding of the cycle-reconstructed photo, which is created using coarse feature maps modulated by $S_{p \rightarrow c}$ and $S_{c \rightarrow p}$ sequentially.

\begin{figure*}
    \centering
    \normalsize
    \begin{tabularx}{\textwidth}{c@{\hspace{4pt}}c@{\hspace{2pt}}c@{\hspace{2pt}}c@{\hspace{2pt}}c}
    \includegraphics[width=0.19\textwidth]{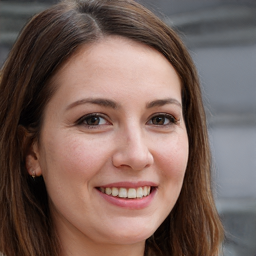} &
    \includegraphics[width=0.19\textwidth]{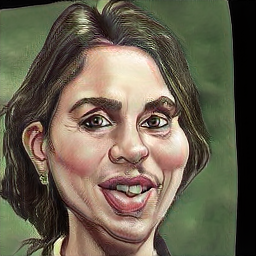} &
    \includegraphics[width=0.19\textwidth]{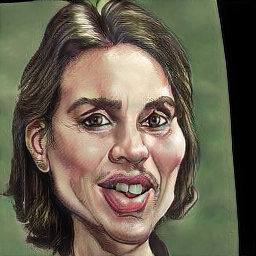} &
    \includegraphics[width=0.19\textwidth]{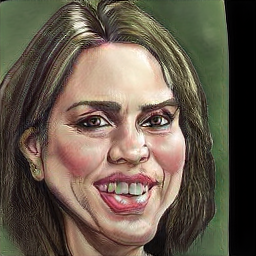} &
    \includegraphics[width=0.19\textwidth]{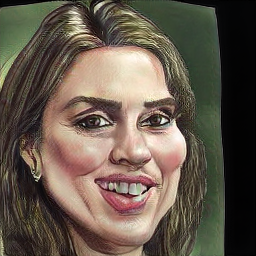} \\
    \includegraphics[width=0.19\textwidth]{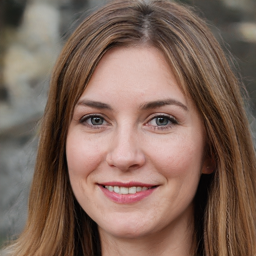} &
    \includegraphics[width=0.19\textwidth]{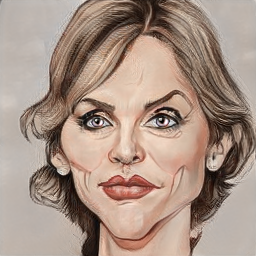} &
    \includegraphics[width=0.19\textwidth]{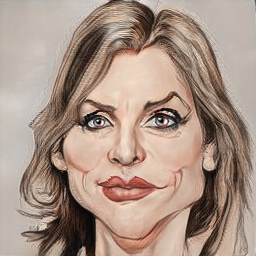} &
    \includegraphics[width=0.19\textwidth]{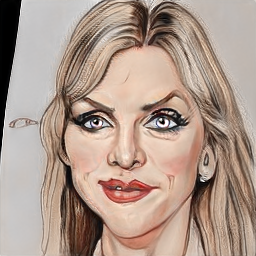} &
    \includegraphics[width=0.19\textwidth]{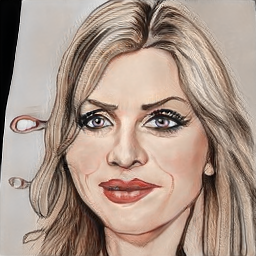} \\
    (a) input &
    (b) $\mathcal{L}_{adv}$ &
    (c) $\mathcal{L}_{adv} + \mathcal{L}_{cyc}$ &
    (d) $\mathcal{L}_{adv} + \mathcal{L}_{attr}$ &
    (e) $\mathcal{L}_{adv} + \mathcal{L}_{cyc} + \mathcal{L}_{attr}$\\
    
    \end{tabularx}
    \vspace{-10pt}
    \caption{\textit{Effect of each loss used for training shape exaggeration blocks.} The input image (a) is transformed to a cartoon-style shape with the effect of adversarial loss $\mathcal{L}_{adv}$ (b). The cartoon-style deformation introduced by $\mathcal{L}_{adv}$ is random and does not respect the input shape. \change{Adding only one of cycle consistency loss $\mathcal{L}_{cyc}$ (c) or attribute matching loss $\mathcal{L}_{attr}$ (d) changes the output to have a more similar facial contour and hair to the input, but the results often show excessive wrinkles or deformed eye shapes. Combining the attribute matching loss $\mathcal{L}_{attr}$ together with the cycle consistency loss $\mathcal{L}_{cyc}$ (e), we obtain caricatures that preserve important features of the input without artifacts.}}
    \label{fig:losseffect}
\end{figure*}

\change{Note that our cycle consistency loss does not require a caricature-to-feature-map encoder because our photo-to-caricature and caricature-to-photo mappings are feature-map-to-feature-map mappings. The mappings do not take images as the input, and we can directly use feature maps to implement the cycle consistency loss.
}

Finally, the cycle consistency loss $\mathcal{L}_{cyc}$ is computed as
\begin{equation}
        \mathcal{L}_{cyc} = \mathcal{L}_{fcyc} + \lambda_{icyc}\mathcal{L}_{icyc}.
\end{equation}

\paragraph{Attribute Matching Loss}
Even with the cycle consistency, the resultant exaggeration is not guaranteed to preserve important features of the input photo. 
To constrain the shape exaggeration blocks to produce valid caricature deformations, we use facial attribute classifiers for photos and caricatures. The facial attribute classifiers are trained on an attribute-labeled dataset, WebCariA \cite{HuoECCV2020WebCariA}. The dataset annotates 50 attributes that describe intrinsic facial shape features such as the sizes of facial components or the overall shape of the face. These attributes are not affected by extrinsic factors such as cosmetics or poses. The annotation is done both for photos and caricatures using the same set of annotation labels.

We constrain p2c-StyleCariGAN to create a caricature with the same attributes as the input photo. We apply similar attribute matching to c2p-StyleCariGAN. The attribute matching loss $\mathcal{L}_{attr}$ is defined using binary cross entropy losses between photo attributes and caricature attributes:
\begin{gather}
\mathcal{L}_{attr}^{p \rightarrow c}=-\mathbb{E}_{w\sim \mathcal{W}}[\phi_{p}(G_{p}(w))\log\phi_{c}(G_{p\rightarrow c}(w))\nonumber\\
+(1 - \phi_{p}(G_{p}(w)))\log(1 - \phi_{c}(G_{p\rightarrow c}(w)))],\nonumber\\[3pt]
\mathcal{L}_{attr}^{c \rightarrow p}=-\mathbb{E}_{w\sim \mathcal{W}}[\phi_{c}(G_{c}(w))\log\phi_{p}(G_{c\rightarrow p}(w))\nonumber\\
+(1 - \phi_{c}(G_{c}(w)))\log(1 - \phi_{p}(G_{c\rightarrow p}(w)))],\nonumber\\[3pt]
\mathcal{L}_{attr} = \mathcal{L}_{attr}^{p \rightarrow c} + \mathcal{L}_{attr}^{c \rightarrow p},
\end{gather}
where $\phi_{p}$ is a photo attribute classifier, $\phi_{c}$ is a caricature attribute classifier, $G_{p}$ is the photo StyleGAN, $G_{c}$ is the caricature StyleGAN, $G_{p \rightarrow c}$ is p2c-StyleCariGAN, and $G_{c \rightarrow p}$ is c2p-StyleCariGAN.

To summarize, our full objective function for training is as follows:
\begin{equation} 
\mathcal{L}_{G}=\lambda_{adv}\mathcal{L}_{adv} + \lambda_{cyc}\mathcal{L}_{cyc} + \lambda_{attr}\mathcal{L}_{attr},
\end{equation} 
where $\lambda_{adv}$, $\lambda_{cyc}$, and $\lambda_{attr}$ are constants defining the loss weights. The effects of different loss components are visualized in \Fig{losseffect}\change{, and more examples can be found in the supplementary material.}

\begin{figure*}[t]
    \centering
    \normalsize
    \begin{tabularx}{\textwidth}{@{\hspace{45pt}}c@{\hspace{0pt}}c@{\hspace{0pt}}c@{\hspace{0pt}}c@{\hspace{2pt}}c}
    \vspace{-3.2pt}
    \includegraphics[width=0.16\textwidth]{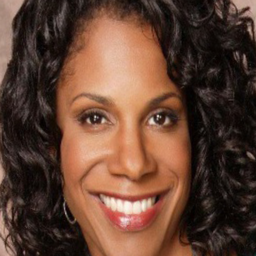} &
    \includegraphics[width=0.16\textwidth]{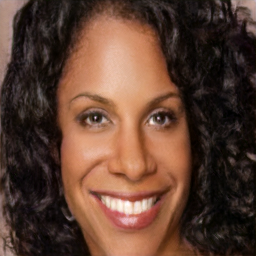} &
    \includegraphics[width=0.16\textwidth]{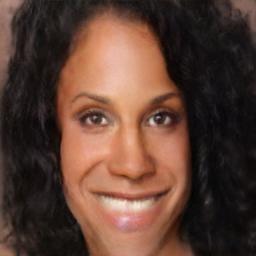} &
    \includegraphics[width=0.16\textwidth]{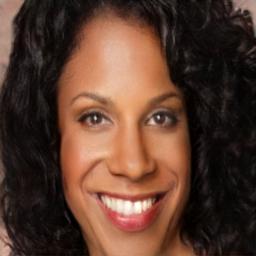} &
    \includegraphics[width=0.16\textwidth]{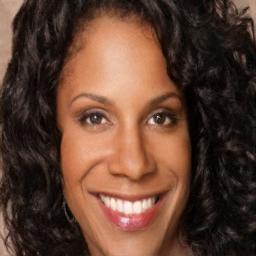} \\
    &
    \includegraphics[width=0.16\textwidth]{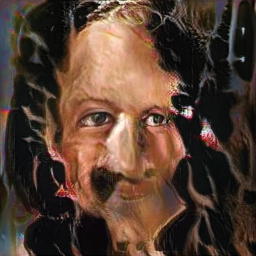} &
    \includegraphics[width=0.16\textwidth]{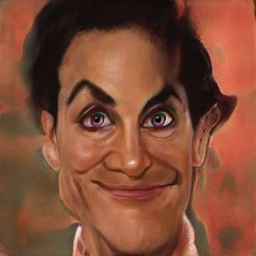} &
    \includegraphics[width=0.16\textwidth]{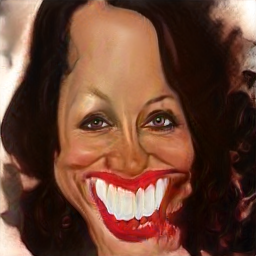} &
    \includegraphics[width=0.16\textwidth]{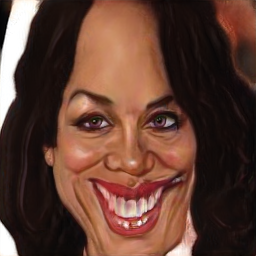} \\
    (a) input &
    (b) Zhu~\Etal  &
    (c) Tewari~\Etal &
    (d) Karras~\Etal &
    (e) Tewari~\Etal\shortcite{tewari2020pie} \\
    &\shortcite{zhu2020domain}&\shortcite{tewari2020pie}&\shortcite{karras2020analyzing}&+ Karras~\Etal\shortcite{karras2020analyzing}
    \end{tabularx}
    \vspace{-8pt}
    \caption{\textit{Comparison of different methods for photo-to-latent embedding.} The input image (a) is embedded into a $\mathcal{W}+$ latent vector using different methods. The reproduced images using the embedded latent vectors are visualized in the top row. The caricatures generated from the embedded vectors using our StyleCariGAN are in the bottom row. 
    All embedded vectors using different methods produce plausible reconstructions, but those from previous methods (b-d) do not result in satisfactory caricatures.
    In contrast, our approach (e) combining \cite{tewari2020pie} and \cite{karras2020analyzing} obtains an embedded vector that achieves a more pleasing caricature. 
    {\footnotesize Input: \copyright Craig Sjodin/Getty Images.}}
    \label{fig:inversion}
    \vspace{-2pt}
\end{figure*}

\subsection{Exaggeration degree control}
Our shape exaggeration blocks enable multi-scale deformation control over the generated images. Since these blocks perform additive modulation on four feature maps representing different spatial scales, we can freely change the amount of modulation by simply attaching a scaling factor to the output of each block, which alters the degree of exaggeration. For example, to reduce the deformation of the overall face shape, we can multiply a weight less than 1 to the output of the first shape exaggeration block before it is added to the original feature map. To remove wrinkles incurred by extreme deformations, we can multiply zero or a small scaling factor to the output of the third or the fourth shape exaggeration block.

\subsection{Photo-to-latent embedding}
\label{sec:method-encoding}

To use a photo image for the input of our StyleCariGAN, 
we optimize the $\mathcal{W}+$ latent code of StyleGAN to reproduce the photo. 
Since there can be multiple 
latent codes that result in similar photos, the mapping from the input photo to the latent code is not unique. However, not all of the possible latent codes are adequate for generating feature maps suitable for image editing. Therefore, we need to choose a good latent code that is meaningful for editing.

The most intuitive solution to obtain the desired latent code is simply applying an existing GAN inversion method. \change{Some GAN inversion methods are good at reconstructing an input image \cite{abdalimage2stylegan, abdal2020image2stylegan++}, but their embedded latent codes are often out-of-distribution \cite{zhu2020domain}. Therefore, we considered GAN inversion methods that work for image editing in the latent space \cite{tewari2020pie, karras2020analyzing}.} 

Hierarchical optimization \cite{tewari2020pie} and StyleGAN2 inversion \cite{karras2020analyzing} showed the most promising results in our task. However, hierarchical optimization, which first finds the latent code in $\mathcal{W}$ space and then updates it in $\mathcal{W}+$ space, is restricted to restoring high-frequency features in images since it does not consider noise optimization. StyleGAN2 inversion optimizes noise as well as latent code, but it may move away from the prior distribution of StyleGAN since it simply initializes the latent code with the mean latent code. To improve existing methods, we combine hierarchical optimization \cite{tewari2020pie} and noise optimization techniques \cite{karras2020analyzing} for latent code embedding.

Specifically, we first find the $\mathcal{W}$ latent code that reconstructs an input facial image via optimization. Then, we optimize $\mathcal{W}+$ latent code and noise simultaneously by initializing the latent code with the obtained $\mathcal{W}$ latent code. We also apply latent code perturbation and noise regularization \cite{karras2020analyzing}. As a result, we can obtain the desired latent code suitable for our framework, as illustrated in \Fig{inversion}.

%% file: tex/4.experiment.tex
\section{Experiments}
\label{sec:experiments}

\begin{figure*}
    \centering
    \normalsize
    \begin{tabularx}{\textwidth}{c@{\hspace{0pt}}c@{\hspace{0pt}}c@{\hspace{0pt}}c@{\hspace{0pt}}c@{\hspace{0pt}}c@{\hspace{0pt}}}
    Input & Ours & U-GAT-IT & StarGAN v2 & WarpGAN & AutoToon \\
    
    \includegraphics[width=0.16\textwidth]{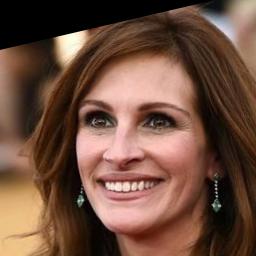} &
    \includegraphics[width=0.16\textwidth]{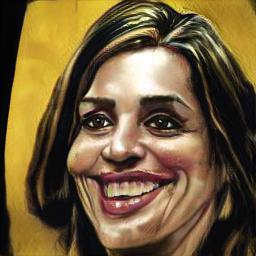} &
    \includegraphics[width=0.16\textwidth]{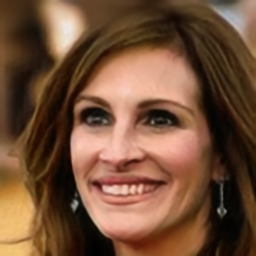} &
    \includegraphics[width=0.16\textwidth]{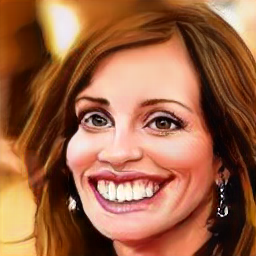} &
    \includegraphics[width=0.16\textwidth]{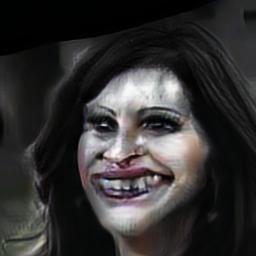} &
    \includegraphics[width=0.16\textwidth]{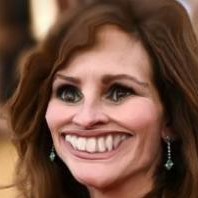} \\
    \includegraphics[width=0.16\textwidth]{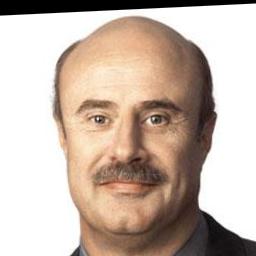} &
    \includegraphics[width=0.16\textwidth]{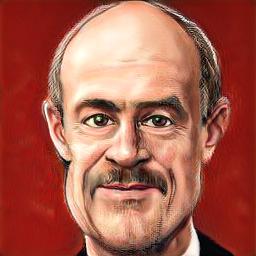} &
    \includegraphics[width=0.16\textwidth]{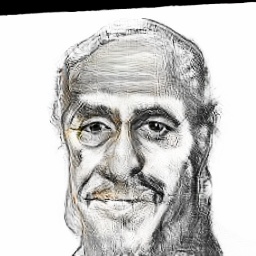} &
    \includegraphics[width=0.16\textwidth]{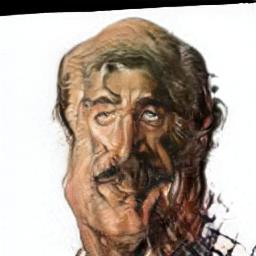} &
    \includegraphics[width=0.16\textwidth]{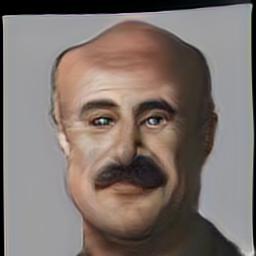} &
    \includegraphics[width=0.16\textwidth]{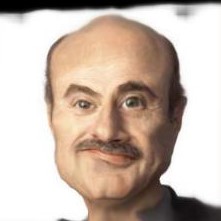} \\
    \includegraphics[width=0.16\textwidth]{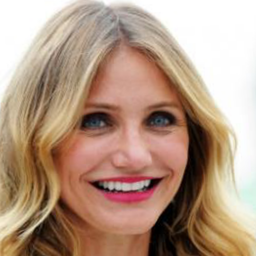} &
    \includegraphics[width=0.16\textwidth]{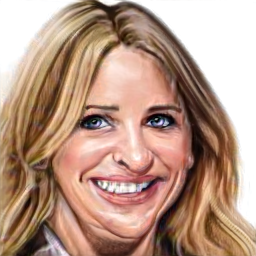} &
    \includegraphics[width=0.16\textwidth]{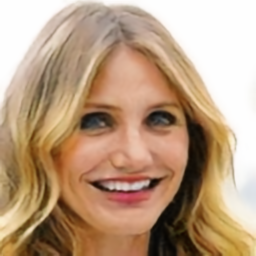} &
    \includegraphics[width=0.16\textwidth]{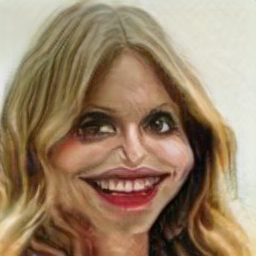} &
    \includegraphics[width=0.16\textwidth]{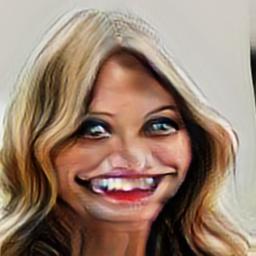} &
    \includegraphics[width=0.16\textwidth]{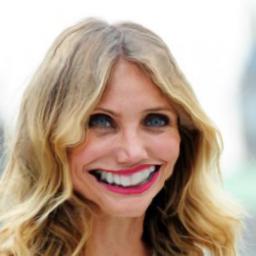} \\
    \includegraphics[width=0.16\textwidth]{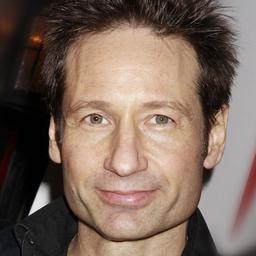} &
    \includegraphics[width=0.16\textwidth]{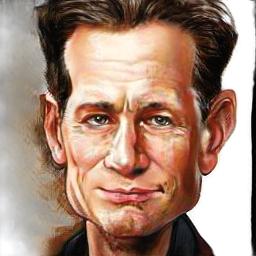} &
    \includegraphics[width=0.16\textwidth]{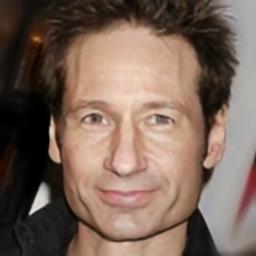} &
    \includegraphics[width=0.16\textwidth]{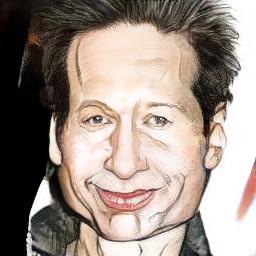} &
    \includegraphics[width=0.16\textwidth]{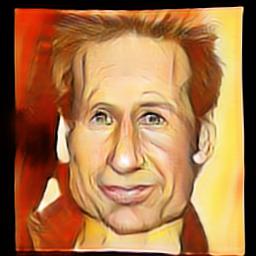} &
    \includegraphics[width=0.16\textwidth]{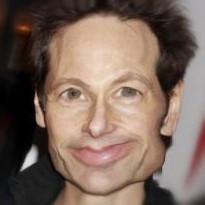} \\
    \includegraphics[width=0.16\textwidth]{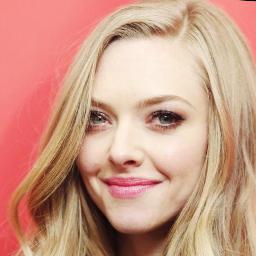} &
    \includegraphics[width=0.16\textwidth]{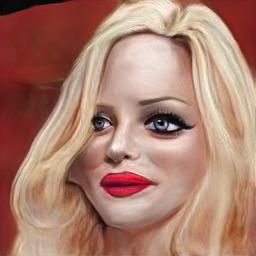} &
    \includegraphics[width=0.16\textwidth]{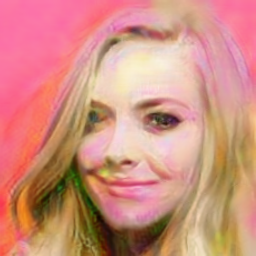} &
    \includegraphics[width=0.16\textwidth]{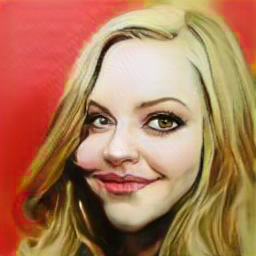} &
    \includegraphics[width=0.16\textwidth]{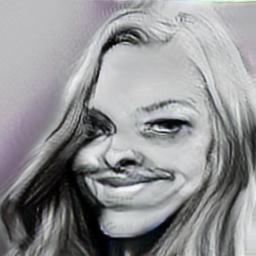} &
    \includegraphics[width=0.16\textwidth]{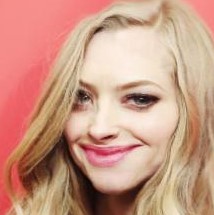} \\
    \includegraphics[width=0.16\textwidth]{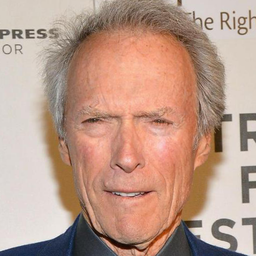} &
    \includegraphics[width=0.16\textwidth]{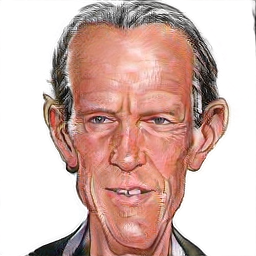} &
    \includegraphics[width=0.16\textwidth]{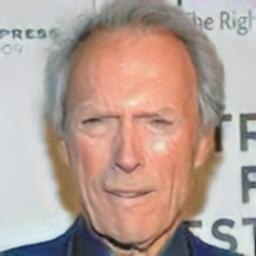} &
    \includegraphics[width=0.16\textwidth]{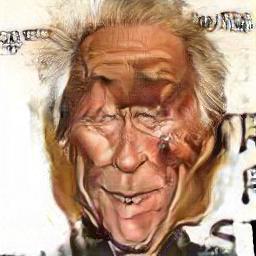} &
    \includegraphics[width=0.16\textwidth]{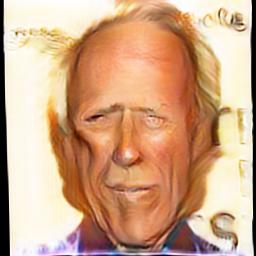} &
    \includegraphics[width=0.16\textwidth]{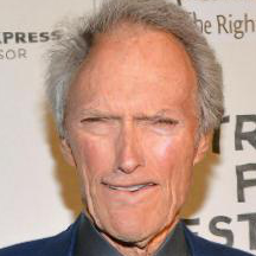} \\
    \end{tabularx}
    \vspace*{-0.4cm}
    \caption{\textit{Comparison with other state-of-the-art methods for image-to-image translation and caricature generation.} Our method can deform and stylize the input faces to produce more realistic and detailed caricatures in comparison to other methods. More visual comparisons can be found in the supplementary material. 
    {\scriptsize Inputs: \copyright Jordan Strauss/Invision/AP, \copyright AF archive/Alamy Stock Photo, \copyright Stuart C. Wilson/Getty Images, \copyright WENN Rights Ltd/Alamy Stock Photo, \copyright Danny Moloshok/Invision/AP, \copyright Slaven Vlasic/Getty Images.}}
    \vspace*{-0.3cm}
    \label{fig:comparison}
\end{figure*}

In this section, we first describe the settings of our experiments. We then evaluate our StyleCariGAN\footnotemark qualitatively and quantitatively and compare it with state-of-the-art methods.
\footnotetext{Our code is available at \url{https://github.com/wonjongg/StyleCariGAN}}

\paragraph{StyleGANs}
Our framework requires two StyleGANs; one generates photos, and the other generates caricatures. Both StyleGANs use the architecture and the training algorithm of StyleGAN2 \cite{karras2020analyzing}. The StyleGAN for photos was trained with FFHQ dataset \cite{karras2019style} resized to $256 \times 256$ resolution. The StyleGAN for caricatures was fine-tuned from the photo model with caricatures in WebCaricature \cite{huo2017webcaricature} using FreezeD \cite{mo2020freeze} and ADA \cite{karras2020training}. The WebCaricature dataset was aligned using five landmarks (the centers of the eyes, the tip of the nose, and the two corners of the mouth) and resized to $256 \times 256$ as well. \change{We used a PyTorch implementation \cite{stylegan2github} for training both models.}

\paragraph{Face embedding network}
To employ the identity cycle consistency loss $\mathcal{L}_{icyc}$, a face embedding network is required. We use a pre-trained FaceNet \cite{schroff2015facenet} as the face embedding network.

\paragraph{Attribute classifiers}
To implement $\mathcal{L}_{attr}$, we need two face attribute classifiers for photos and caricatures, respectively. We train the attribute classifiers using WebCariA dataset \cite{HuoECCV2020WebCariA}\change{, which provides labels for both photos and caricatures. We found that the label distributions for photos and caricatures are reasonably similar in the dataset.} The backbone architecture of the attribute classifiers is ResNet-18 \cite{he2016identity} with the only change in the output channel size of the last fully connected layer. Our output channel size is $50$, which is the number of attributes in WebCariA. We fine-tuned the pre-trained ResNet-18 provided by PyTorch \cite{pytorch}. The test accuracy on the test set of the WebCariA dataset was $85\%$ for photos and $82\%$ for caricatures.

\paragraph{Training shape exaggeration blocks}
\change{A shape exaggeration block consists of two convolutional layers, each with a leaky ReLU. Each convolutional layer has kernel size = 3, stride = 1, and padding = 1. The leaky ReLU layers have a negative slope 0.2.}

We use the Adam optimizer \change{\cite{kingma2014adam}} in PyTorch with $\beta_{1} = 0$, $\beta_{2} = 0.99$, and learning rate $0.002$. Each mini-batch with a batch size 32 consists of a randomly generated photos and caricatures. \change{We stopped training after $1,000$ iterations}. We empirically set the weights for the losses as $\lambda_{adv} = 1$, $\lambda_{GAN} = 10$, $\lambda_{cyc} = 10$, $\lambda_{icyc} = 1000$, and $\lambda_{attribute} = 10$. The training time with 4 NVIDIA Quadro RTX 8000 (48 GB) GPUs was about 8 hours.

\change{
\paragraph{Running time}
In test time, the GAN inversion from an input photo to the latent code takes three to four minutes. After the inversion, generating a caricature image takes about 40 ms. The measurement was done on a server with NVIDIA Quadro RTX 8000 and Intel Xeon Gold 6226R. \blue{We use $256 \times 256$ size for both the input photo and the output caricature.}}

\subsection{Comparison to state-of-the-art methods}
We qualitatively evaluate our method by comparing with two classes of methods: generic image-to-image translation methods and deep caricature generation methods (\Fig{comparison}).

\begin{figure*}[t]
    \centering
    \normalsize
    \begin{tabularx}{\textwidth}{c@{\hspace{0pt}}c@{\hspace{0pt}}c}
    \vspace{-3.2pt}
    \includegraphics[width=0.33\textwidth]{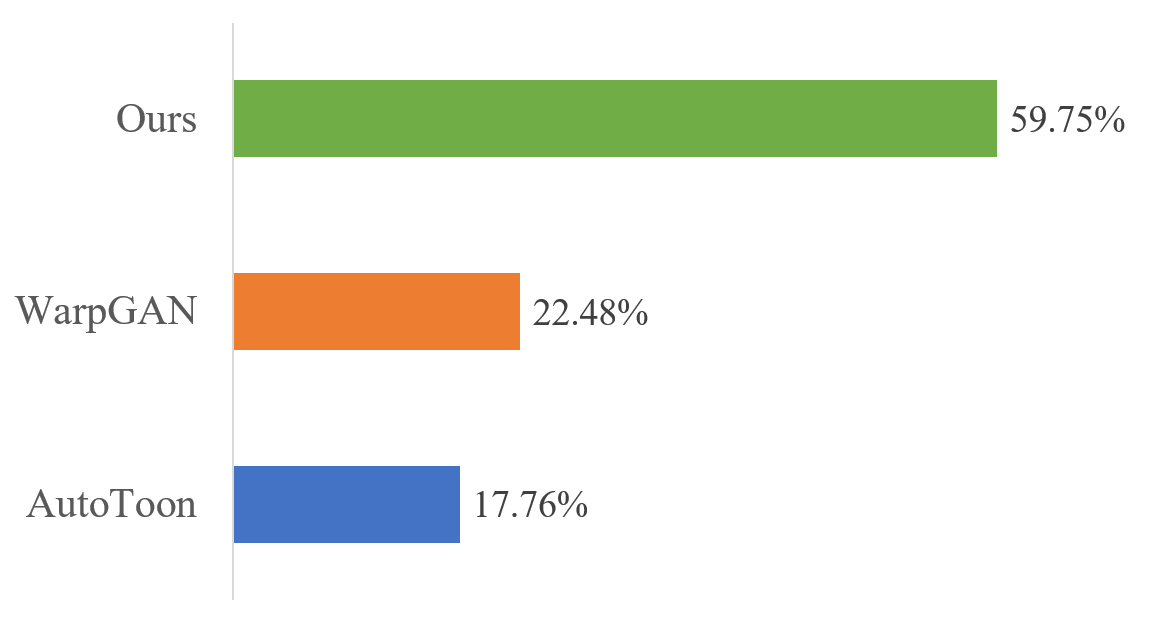} &
    \includegraphics[width=0.33\textwidth]{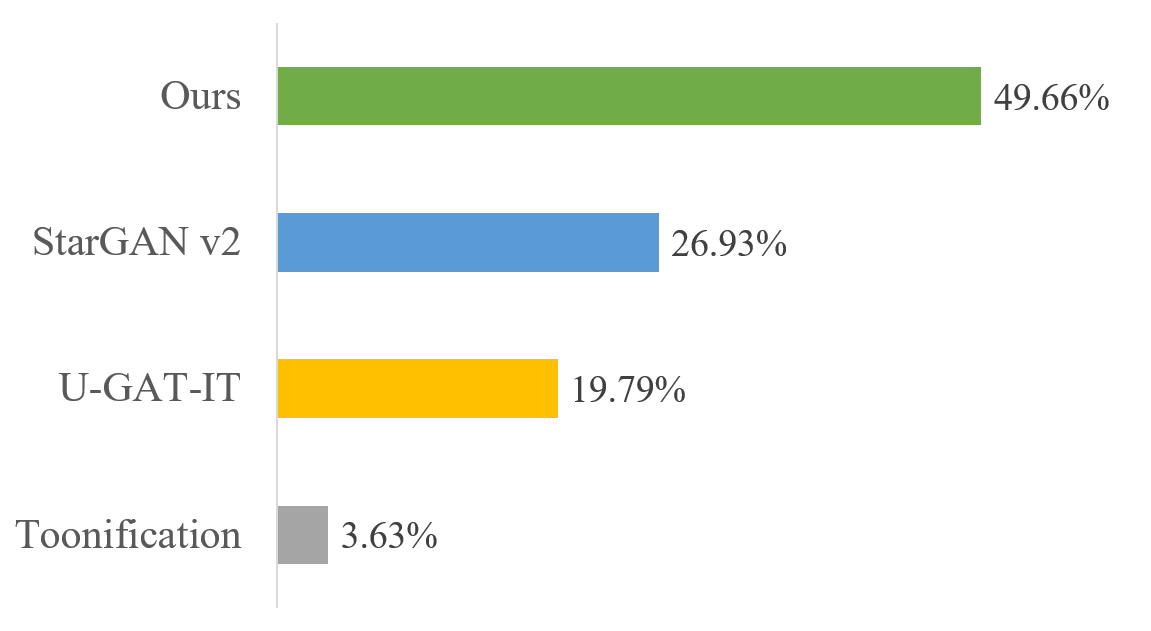} &
    \includegraphics[width=0.33\textwidth]{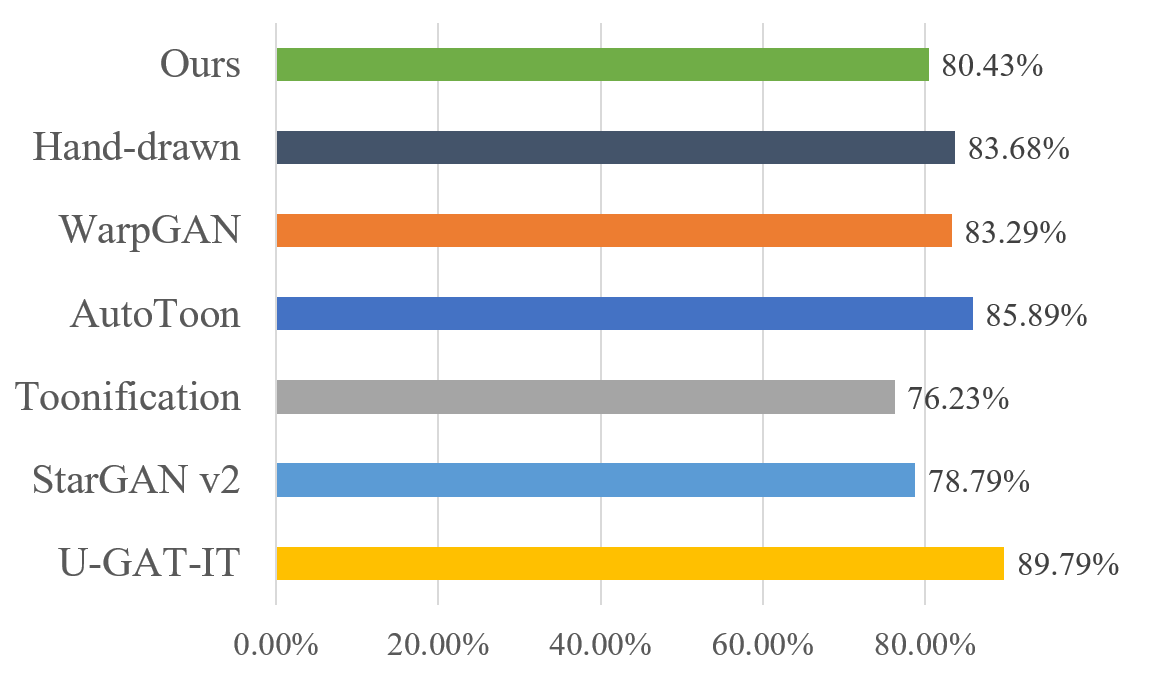} \\
    \hdashline[1.0pt/2pt]
    \includegraphics[width=0.33\textwidth]{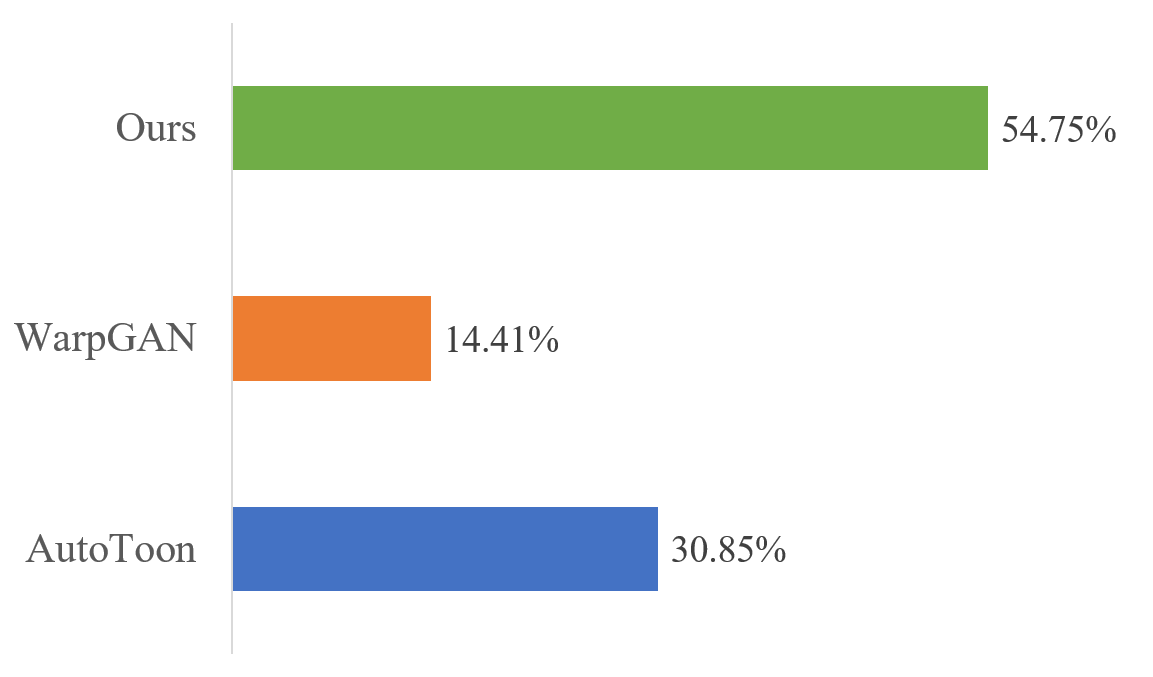} &
    \includegraphics[width=0.33\textwidth]{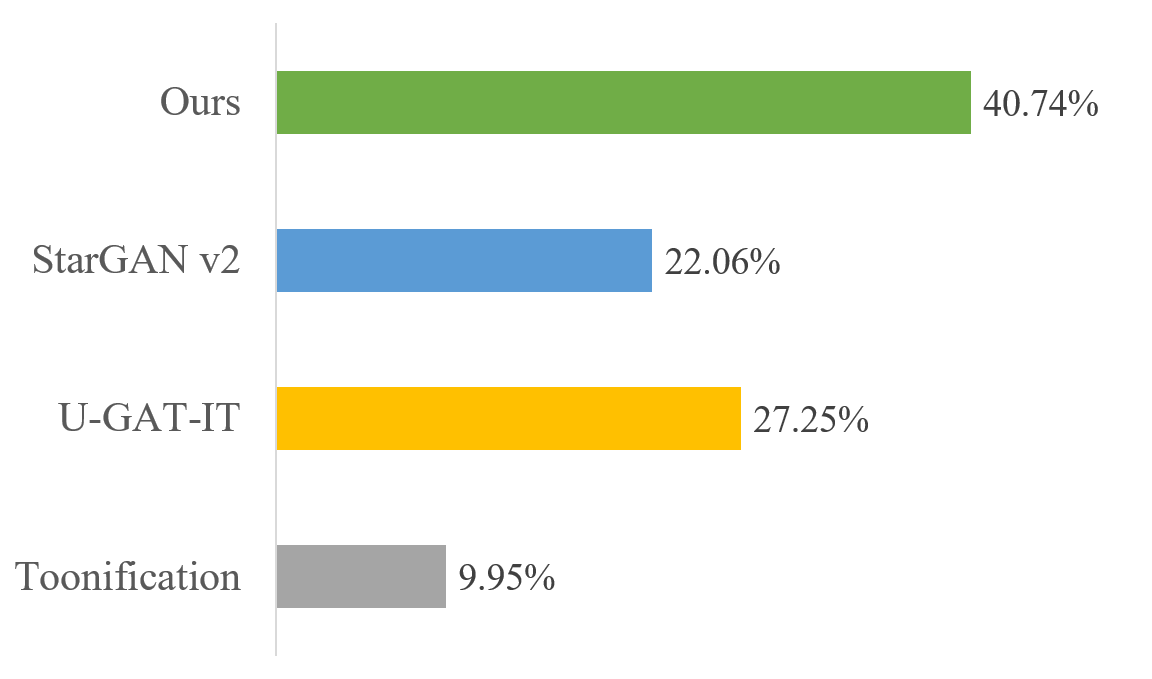} &
    \includegraphics[width=0.33\textwidth]{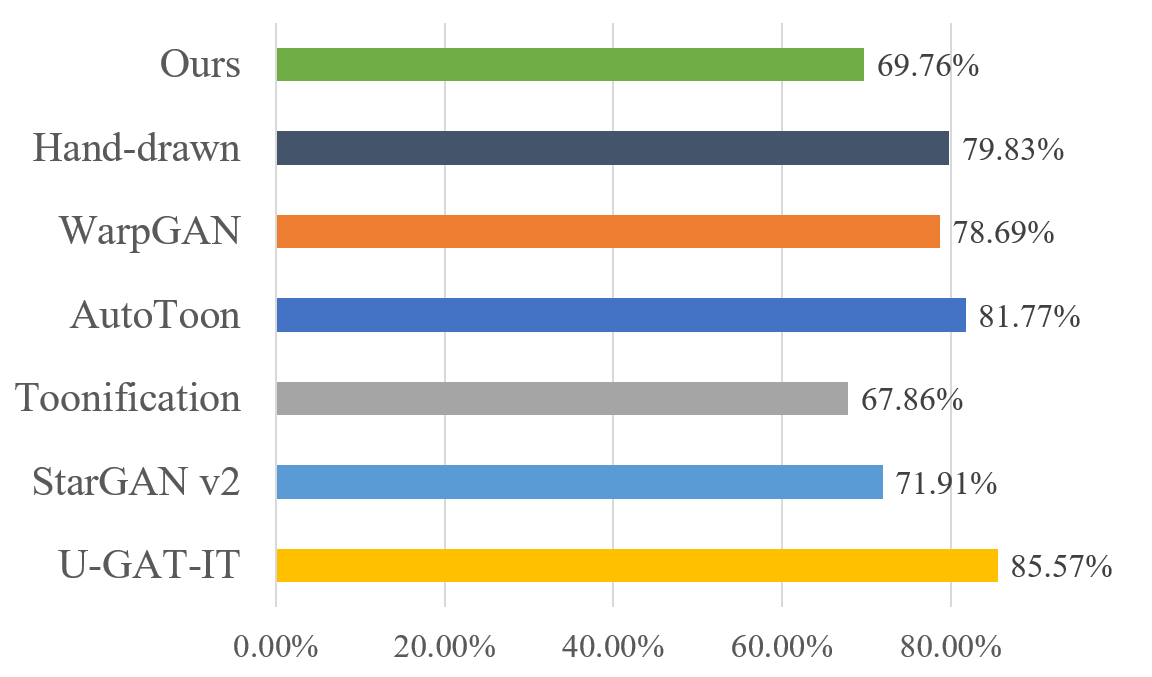} \\
    (a) &
    (b) &
    (c) \\
    \end{tabularx}
    \vspace{-8pt}
    \caption{\change{\textit{User study results.} Top and bottom parts show the results when our exaggeration factors are 0.5 and 1.0, respectively. (a, b) Faithfulness to hand-drawn caricature styles. Our method outperforms other methods with a large margin in terms of the quality of caricature styles. (c) Identity preservation. Our method shows comparable identification rates to other methods.}}
    \label{fig:userstudy}
    \vspace{-1pt}
\end{figure*}

We compare our results with U-GAT-IT \cite{kim2019u} and StarGAN v2 \cite{choi2020stargan}. Among various image-to-image translation models, the two models have showcased visually pleasing results in their work when translating two domains with large shape changes. We trained U-GAT-IT and StarGAN v2 from scratch on the WebCaricature dataset using their official implementations. 
U-GAT-IT does not successfully generate caricatures from the input, creating blurry images. Sometimes the changes in the output compared to the input are minuscule. StarGAN v2 generates more stable results, but sometimes artifacts are found in textures. Besides, the degree of exaggeration is smaller than ours.

We also compare our results with WarpGAN \cite{shi2019warpgan} and AutoToon \cite{gong2020autotoon}, which are state-of-the-art methods for caricature generation based on deep learning. We used the pre-trained models of WarpGAN and AutoToon released by the authors. Similar to U-GAT-IT and StarGAN v2, WarpGAN was trained on the WebCaricature dataset. In contrast, AutoToon was trained on its own dataset containing photo and caricature pairs for supervised learning. We also present comparisons with CariGANs \cite{cao2018carigans} \change{and CariGAN \cite{li2020carigan}} in the supplementary material.

As can be seen from \Fig{comparison}, these deep learning based caricature generation methods afford large shape changes using explicit 2D image deformations. However, the generated shape deformations do not provide enough details to create realistic caricatures. WarpGAN uses only 16 sparse control points to handle deformations, and AutoToon is trained with only 101 photo-caricature pairs. Because of the limited density of control points and the sparsity of training samples, the generalization of these methods to realistic caricatures would be hard. Visually inspected, the results of our method are not only more pleasing but also closer to artist-created contents.

\change{
\subsection{Quantitative analysis}
We quantitatively measured the faithfulness to caricature image distribution with an evaluation of FID \cite{heusel2017gans}. Compared to other state-of-the-art methods (AutoToon, U-GAT-IT, WarpGAN, StarGAN v2), our method showed the best FID score (\Tbl{FID}). This shows that our generated caricatures are closest to the distribution of caricatures. We trained U-GAT-IT, StarGAN v2, and ours with FFHQ \cite{karras2019style} for photo and WebCaricature \cite{huo2017webcaricature} for caricature dataset. We used officially released codes and models for AutoToon and WarpGAN. The reported values are FIDs between the generated caricatures and all caricatures in the WebCaricature dataset. The generated caricatures are created from all images in the CelebA \cite{liu2015faceattributes} dataset, which was not involved in training.}

\begin{table}[t]
\caption{\textit{FIDs between generated and hand-drawn caricatures.} Lower is better.}
\vspace*{-0.2cm}
\label{tbl:FID}
    \centering
    \small
    \begin{tabularx}{\columnwidth}{Y Y}
\toprule
\textbf{Method} & \textbf{FID}
\\
\midrule
AutoToon & 114.84\\
U-GAT-IT & 92.79\\
WarpGAN & 74.60\\
StarGAN v2 & 57.94 \\
Ours & \textbf{52.35}\\
\bottomrule
\end{tabularx}
\vspace*{-0.3cm}
\end{table}

\change{
\subsection{Perceptual study}
We perceptually evaluated the faithfulness to hand-drawn caricature styles through a user study. Given a photo, we asked users to select the best caricature that looks like a hand-drawn caricature among different caricatures generated from state-of-the-arts methods and ours. We compared the responses for WarpGAN, AutoToon, U-GAT-IT, StarGANv2, and Toonification with ours. We evaluate our method using two different exaggeration factors (0.5 and 1.0). We run the experiment twice, one for each exaggeration factor of our method. Different users were involved in each experiment to prevent a learning effect.
There are several methods to compare, and it could be hard for the user to answer consistently if the caricature results from all methods are shown together for each question.
Therefore, we run the user study by splitting the methods into two groups: specifically designed caricature generator methods (WarpGAN, AutoToon) and general image-to-image translation methods (U-GAT-IT, StarGANv2, Toonification). 

For each group, each user was asked 30 questions randomly sampled from the pool of 71 questions. 
The question pool was constructed by generating caricatures for the input photos previously used for the compared methods and the input photos in \Fig{comparison}.
We added five duplicate questions to filter out random selections. We excluded users that show inconsistent answers in more than three duplicate questions to obtain valid responses from 60 users for each group. Before starting the experiment, we asked each user five training questions using hand-drawn caricatures to expose the user to realistic caricature styles. In each training question, a user was asked to pick the best hand-drawn caricature that matches the input photo. The user study was done using Amazon Mechanical Turk. In both of the groups, regardless of the exaggeration factor, our method significantly outperformed the previous methods (\Figs{userstudy}a-b).
The questions and results are included in the supplementary material.}

\change{We also perceptually evaluated the degree of identity preservation of our method through another user study.
Similarly to \cite{cao2018carigans},
we asked users to pick the photo that matches the identity of an input caricature. In each question, the input caricature was one of Hand-drawn, WarpGAN, AutoToon, Toonification, U-GAT-IT, StarGANv2, and ours. As in the other user study, we evaluated our method using two different exaggeration factors (0.5 and 1.0). For a fair comparison, we run the experiment twice with different exaggeration factors of our method, instead of running the experiment once with both results of ours, to balance the number of exposures per method. Consequently, in each experiment, there are seven caricature generators in total. 

We provided five photo choices per question: one containing the answer with the same identity as the input for the caricature but with a different pose, and the rest containing similar faces to the answer. The similar faces for the wrong choices were selected by first encoding the answer into a FaceNet \cite{schroff2015facenet} feature and estimating pose using \cite{deng2019accurate},
then selecting images that show close feature distances and similar poses to the answer among the union of FFHQ and CelebA datasets. 

We run each experiment on caricatures of 34 randomly selected identities from WebCaricature \cite{huo2017webcaricature}. With the seven methods to evaluate, there are $34 \times 7$ caricatures. We created seven sets of questions, each set containing 39 questions that consist of 34 questions and five duplicate questions for filtering out randomly-picking users. The caricatures of the sets are mutually exclusive and the union of the sets covers all the $34 \times 7$ caricatures. For each set, we asked 50 users to answer the questions. Before starting the session, we asked each user five training questions, where the inputs are hand-drawn caricatures and the correct answers are shown to the user. The user study was done using Amazon Mechanical Turk. The results show that ours (0.5) shows comparable identity preservation to other methods including hand-drawn caricatures, while ours (1.0) shows lower identity preservation (\Fig{userstudy}c). We included the questions and responses in the supplementary material.}

\change{Some baselines such as AutoToon, WarpGAN, and U-GAT-IT show better identity preservation than hand-drawn caricatures. 
The implementation of AutoToon available on \blue{the internet} and we used for experiments only deforms the input image without changing the textures, leaving clues for inferring the identity. WarpGAN is trained with a reconstruction loss that forces the output to be similar to the input, 
preserving hints for inferring the input identity. U-GAT-IT often generates results that are not very different from the input.
Our method generates caricatures that are close to hand-drawn caricatures, and with a proper exaggeration control, can generate caricatures with comparable identity preservation to hand-drawings.
In \Fig{userstudy}c, the results for Hand-drawn and five previous methods differ among the two experiments on top and bottom, because the users participated in the Amazon Mechanical Turk experiments were not the same although the same sets of caricatures were used.
}

\begin{table}[t]
\caption{\textit{Quantitative analysis on attribute preservation.} The average accuracy of all identity-matching photo and hand-drawn caricature pairs is the upper bound, and the average accuracy of randomly selected pairs is the lower bound. For the identity-matching hand-drawn case, ground-truth labels are annotations in the dataset, and predicted labels are from our attribute classifiers. StyleCariGAN shows a reasonably high average accuracy compared to hand-drawn caricatures with predicted labels.}
\label{tbl:attribute-preservation}
\vspace*{-0.2cm}
    \centering
    \small
    \begin{tabularx}{\columnwidth}{>{\centering}X Y}
\toprule
\textbf{Caricature source} & \textbf{Accuracy}
\\
\midrule
identity-matching hand-drawn (using GT labels)& 85.26\%
\\
\vspace{0.1cm}
identity-matching hand-drawn (using predicted labels)& 73.73\%
\\
\vspace{0.1cm}
StyleCariGAN \hspace{1.5cm} (using predicted labels)& 70.38\%
\\
\vspace{0.1cm}
random hand-drawn \hspace{1.5cm} (using GT labels)& 66.16\%
\\
\bottomrule
\end{tabularx}
\vspace*{-0.2cm}
\end{table}

\subsection{Attribute preservation accuracy}

To evaluate the attribute preservation capability of our StyleCariGAN, we analyze the detected attributes of our caricature results. Given a pair of photo attributes and caricature attributes, we calculate the matching accuracy as $m / M$, where $m$ is the number of matched elements between photo and caricature attributes, and $M$ is the total number of attributes. We first compute the upper and lower bounds of the accuracy using hand-drawn caricatures, then show our level of attribute preservation by comparing our accuracy to the bounds.

We use WebCariA \cite{HuoECCV2020WebCariA} dataset for the evaluation. The dataset provides labels for photos and hand-drawn caricatures. An attribute classifier is pre-trained on the caricatures of the WebCariA dataset to detect attributes from caricature images. For our evaluation, we filter the attribute labels of each photo by marking only the labels that are shared at least by 50\% of all identity-matching photos. The filtering is done to remove noisy labels coming from extrinsic factors such as pose. 

In this experiment, the upper bound is the average accuracy of all identity-matching photo-caricature pairs in the dataset, while the lower bound is the average accuracy of randomly-selected photo-caricature pairs. Our accuracy is calculated using StyleCariGAN results generated from all photos in the dataset. The evaluation result shows that our method preserves facial attributes reasonably well in the generated caricatures (\Tbl{attribute-preservation}). \change{Note that WebCariA dataset has many mutually exclusive labels, which lead to many zeros in the attribute vectors. Consequently, attribute vectors could be similar even for random pairs, and the accuracy for random hand-drawn becomes relatively high.}

\begin{figure}[t]
    \centering
    \normalsize
    \begin{tabularx}{\columnwidth}{c@{\hspace{3pt}}c@{\hspace{0pt}}c@{\hspace{0pt}}c}
    \includegraphics[align=c,width=0.32\columnwidth]{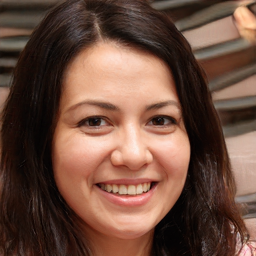} &
     \includegraphics[align=c,width=0.32\columnwidth]{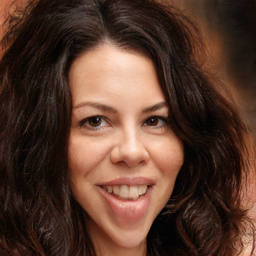} & 
     \includegraphics[align=c,width=0.32\columnwidth]{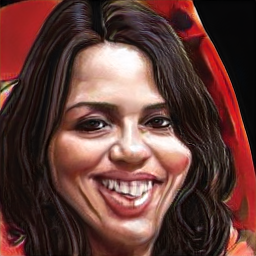}\\
    \includegraphics[align=c,width=0.32\columnwidth]{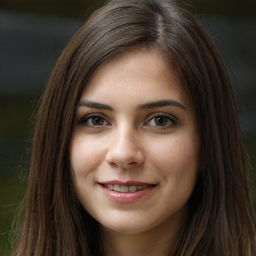} &
    \includegraphics[align=c,width=0.32\columnwidth]{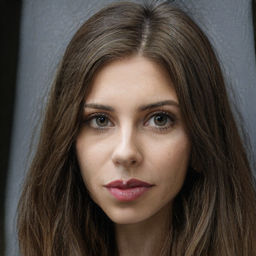} &
    \includegraphics[align=c,width=0.32\columnwidth]{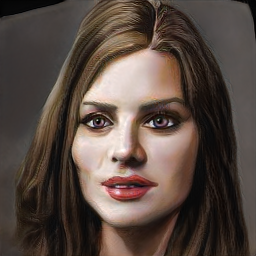} \\
    (a) &
    (b) &
    (c) \\
    \end{tabularx}
    \vspace{-0.2cm}
    \caption{\textit{Comparison with Toonification~\cite{pinkney2020blending}.} Toonification (b) blends an arbitrary caricature structure that largely changes the characteristics of the input (a). For example, Toonification does not preserve the overall facial shape of the input. Our framework creates shape deformations from the input image using \textit{shape exaggeration blocks} trained explicitly to generate caricatures that preserves important visual features. As a result, our framework generates caricatures (c) that possesses exaggerated input structures and cartoon-style rendering.}
    \vspace{-0.1cm}
    \label{fig:toonification}
    \vspace{-0.3cm}
\end{figure}

\subsection{Comparison to \textit{Toonification} based on StyleGAN}
\label{sec:toonification}

Pinkney~and~Adler~\shortcite{pinkney2020blending} proposed the \textit{Toonification} method that adds a cartoon structure into the input photo using StyleGAN layer swapping. The method generates a photo-realistic rendering on top of the cartoon structure by mixing coarse feature maps of cartoons and fine feature maps of photos. Although Toonification and our framework share the same idea of layer swapping, the goals and the results of the two frameworks are different. Toonification simply performs layer swapping using two StyleGANs trained for photos and cartoon images to create a cartoon image from an input photo. Our framework also performs layer swapping but our shape exaggeration blocks additionally create meaningful and facial-attribute-preserving shape deformations from the input photo. \Fig{toonification} shows a visual comparison, where our caricature StyleGAN is used for Toonification, instead of a StyleGAN trained to generate cartoon images. In the comparison, Toonification simply blends arbitrary caricature-style structures into the input photos. In contrast, our framework generates caricatures that not only have pleasing caricature styles but also preserve important features of the input photos. 

\change{
Pinkney~and~Adler~\shortcite{pinkney2020blending} also mentioned that an image with cartoon-like textures can be created by using photo-StyleGAN features as the coarse layers and caricature-StyleGAN features as the fine layers.
In that case, the method becomes identical to our setting except there are no shape exaggeration blocks. The results of this case using different mixing boundaries can be observed in \Fig{layermix}. However, the deformations obtained by simple layer swapping are hard to control for achieving desirable caricature styles. For example, a result generated with more caricature layers can map a photo of a woman to a caricature image of an old man as seen in \Fig{layermix}. On the other hand, if we take few caricature layers, we cannot obtain large enough deformations. Instead of inducing deformations with layer swapping only, we create caricature deformations by using shape exaggeration blocks that are directly supervised with losses designed for training a caricature generator.}

\begin{figure}
    \centering
    \normalsize
    \begin{tabularx}{\columnwidth}{c@{\hspace{3pt}}c@{\hspace{3pt}}c@{\hspace{3pt}}c}
    \vspace{-1.2pt}
    \rotatebox[origin=c]{90}{Input caricature} &
    \includegraphics[align=c, width=0.3\columnwidth]{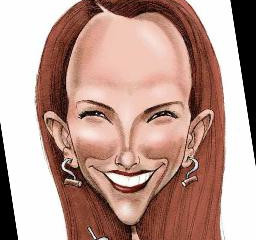} &
    \includegraphics[align=c, width=0.3\columnwidth]{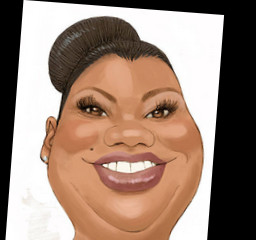} &
    \includegraphics[align=c, width=0.3\columnwidth]{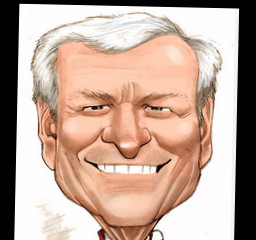} \\
    \rotatebox[origin=c]{90}{Output photo} &
    \includegraphics[align=c, width=0.3\columnwidth]{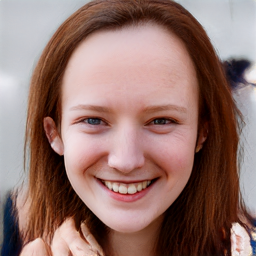} &
    \includegraphics[align=c, width=0.3\columnwidth]{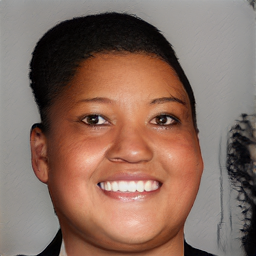} &
    \includegraphics[align=c, width=0.3\columnwidth]{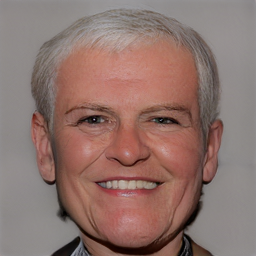} \\
    \end{tabularx}
    \vspace{-0.1cm}
    \caption{\textit{Caricature-to-photo translation.} Given caricatures, we can generate photos using the inverse mapping of StyleCariGAN trained for cycle consistency. The generated photos share important visual features with the input caricatures.
    {\footnotesize Input caricatures: \copyright MCT/Getty Images.}}
    \vspace{-0.1cm}
    \label{fig:cari2photo}
    \vspace{-0.3cm}
\end{figure}

\begin{figure*}[t]
    \centering
    \normalsize
    \begin{tabularx}{\textwidth}{@{\hspace{35pt}}c@{\hspace{5pt}}c@{\hspace{0pt}}c@{\hspace{0pt}}c@{\hspace{0pt}}c@{\hspace{0pt}}c}
    & 0.0 & 0.25 & 0.5 & 0.75 & 1.0 \\
    \vspace{-3.3pt}
    $\alpha_{1, 2, 3, 4}$&
    \includegraphics[align=c, width=0.155\textwidth]{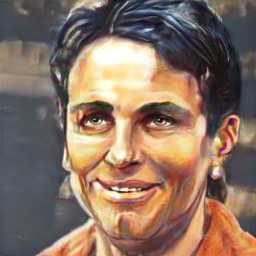} &
    \includegraphics[align=c, width=0.155\textwidth]{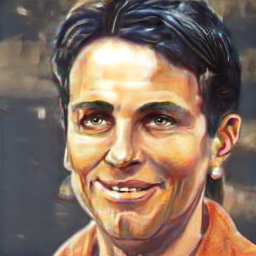} &
    \includegraphics[align=c, width=0.155\textwidth]{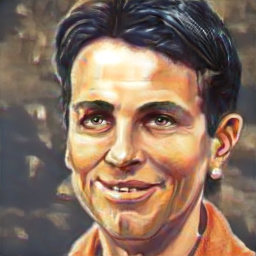} &
    \includegraphics[align=c, width=0.155\textwidth]{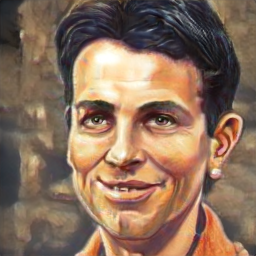} &
    \includegraphics[align=c, width=0.155\textwidth]{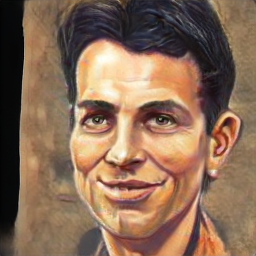} \\
    \vspace{-3.3pt}
    $\alpha_{1, 2, 3}$&
    \includegraphics[align=c, width=0.155\textwidth]{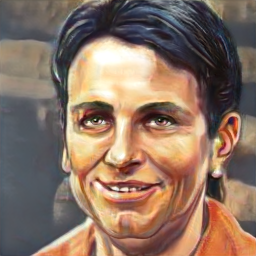} &
    \includegraphics[align=c, width=0.155\textwidth]{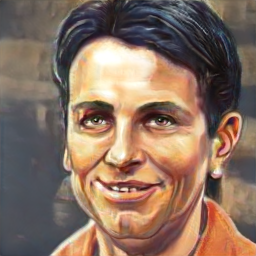} &
    \includegraphics[align=c, width=0.155\textwidth]{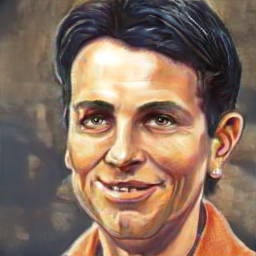} &
    \includegraphics[align=c, width=0.155\textwidth]{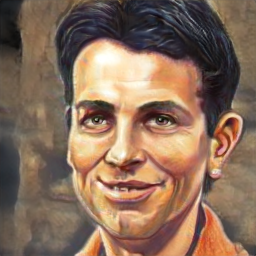} &
    \includegraphics[align=c, width=0.155\textwidth]{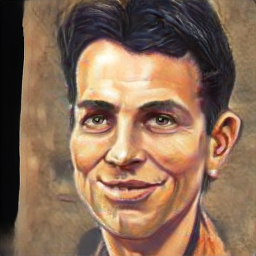} \\
    \vspace{-3.3pt}
    $\alpha_{1, 2}$&
    \includegraphics[align=c, width=0.155\textwidth]{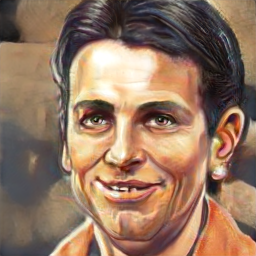} &
    \includegraphics[align=c, width=0.155\textwidth]{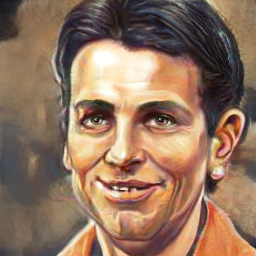} &
    \includegraphics[align=c, width=0.155\textwidth]{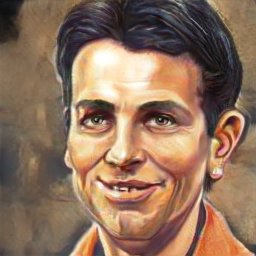} &
    \includegraphics[align=c, width=0.155\textwidth]{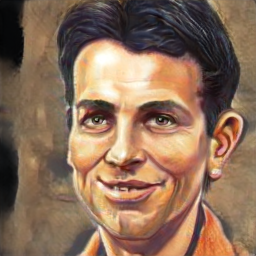} &
    \includegraphics[align=c, width=0.155\textwidth]{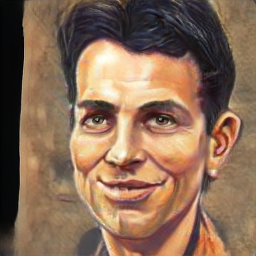} \\
    $\alpha_{1}$&
    \includegraphics[align=c, width=0.155\textwidth]{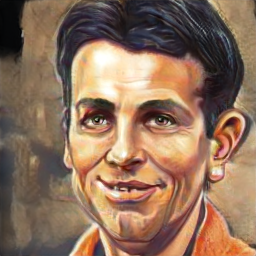} &
    \includegraphics[align=c, width=0.155\textwidth]{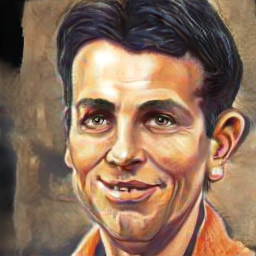} &
    \includegraphics[align=c, width=0.155\textwidth]{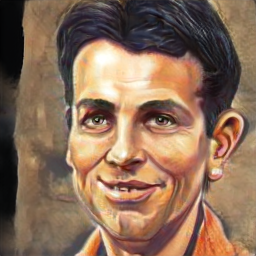} &
    \includegraphics[align=c, width=0.155\textwidth]{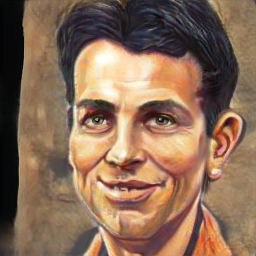} &
    \includegraphics[align=c, width=0.155\textwidth]{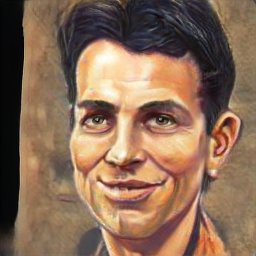} \\
    \end{tabularx}
    \vspace{-0.1cm}
    \caption{\textit{Multi-scale control over shape exaggerations.} StyleCariGAN provides 4-scale control for exaggeration degrees. $\alpha_1$ controls the exaggeration of the coarsest scale, and $\alpha_4$ controls the finest scale. Each row shows different scales of edits. \change{At each row, the set of parameters in the label is modulated, while unspecified parameters are fixed to 1. The first row shows the effects of shape exaggeration blocks, from no exaggeration to full exaggeration. Removing the blocks results in a caricature with photo-like structures (Row 1, Column 1). Applying all the blocks results in a caricature with desirable and realistic deformations (Row 1, Column 4). More examples with/without shape exaggeration blocks can be found in the supplementary material. The other rows show multi-scale control of exaggerations. Adding feature modulations of different scales from $\alpha_4$ to $\alpha_2$ produces structure deformations of increasing scales (Rows 2-4, Column 1). Parameter $\alpha_1$ for the coarsest feature modulation affects overall facial shape (Row 4, Columns 1-4). Note that the caricature results in the rightmost column are all the same since their parameter values are all 1.} 
    }
    \label{fig:app2}
\end{figure*}

\subsection{Caricature-to-photo translation}

Since StyleCariGAN is trained with cycle consistency, it has an inverse mapping for caricature-to-photo translation. To translate a caricature to a photo, we first encode a caricature image into the $\mathcal{W}+$ space of the caricature StyleGAN. Given the encoded vector as the input, we simply use \textit{c2p-StyleCariGAN} trained for cycle consistency to generate a photo that resembles the given caricature. As illustrated in \Fig{cari2photo}, our method can generate convincing caricature-to-photo translation results.

\subsection{Exaggeration control}
\change{The first row of \Fig{app2} demonstrates the contribution of shape exaggeration blocks, where the leftmost and rightmost images show the results without/with the blocks, respectively. It clearly shows our shape exaggeration blocks play a critical role for generating demanded shape changes in caricatures while preserving overall color stylization.
\Fig{app2} also shows some typical examples of controlling the degree of exaggerations in the caricature results by attaching scaling factors to shape exaggeration blocks.
More examples without/with shape exaggeration blocks can be found in the supplementary material.}

\subsection{StyleGAN manipulation}
\label{sec:manipulation}
Since our framework uses StyleGAN as the backbone network, various kinds of StyleGAN manipulations can be applied, including caricature color style customization and caricature expression control. Note that our framework does not require additional training process for such manipulations. 

\begin{figure}[h]
    \centering
    \includegraphics[width=0.48\textwidth]{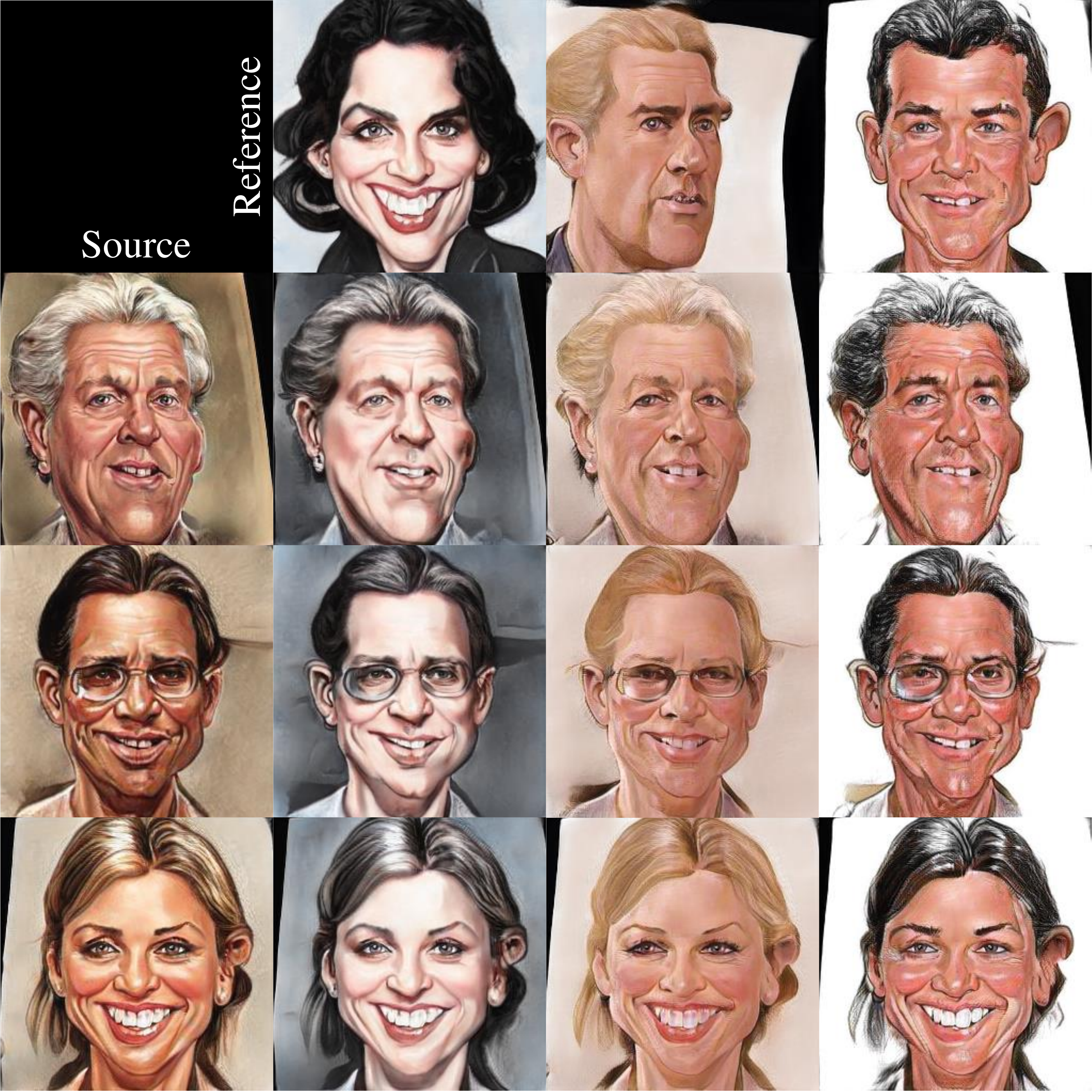}
    \caption{\textit{Caricature appearance style selection.} Our framework can apply various appearance styles to the generated caricature shapes using a set of curated reference caricatures.}
    \label{fig:app1}
    \vspace{-0.6cm}
\end{figure}

\paragraph{Caricature appearance style selection}
In caricature generation, providing user control on appearance style is desirable. We can build such a control with style mixing \cite{karras2019style}, which can apply the tone and texture of a reference image onto the structure of a source image by injecting fine-scale latent codes of the reference image to the latent code of the source image. We use $\mathcal{W}+$ latent codes to perform the style mixing. A set of reference appearances are curated from randomly generated caricature samples using the caricature StyleGAN.
As shown in \Fig{app1}, the appearance style can be flexibly changed with different reference styles, while the original structural contents can be well preserved.

\paragraph{Caricature expression manipulation}
Our framework can be readily used with other StyleGAN-based semantic image editing methods. Recent studies \cite{shen2020closed, shen2020interfacegan} have proposed semantic face editing based on a pretrained GAN. The basic idea of those methods is to find latent-space directions corresponding to the semantics, \emph{e.g.}, smile, and then move the latent code of an input image along the desired directions. In \Fig{app3}, we use InterFaceGAN \cite{shen2020interfacegan} to search for the direction of editing images towards smiling expression. The direction search is performed using the plain StyleGAN for photos. 
Then, we edit the latent code of the input image along the direction to modulate the magnitude of smile. StyleCariGAN produces a caricature image with the modulated smile magnitude by taking the edited latent code as input.
As shown in the examples, this simple approach can add strong cartoon-like smiles without loss of identity information.

%% file: tex/5.conclusion.tex
\section{Conclusion}

In this paper, we presented StyleCariGAN, a novel caricature generation framework based on StyleGAN. Our framework handles the problem of caricature generation by modulating coarse-level feature maps with shape exaggeration blocks and swapping fine-level layers to the corresponding layers of the StyleGAN trained for caricature images.
The shape exaggeration blocks are supervised to produce feature modulations for realistic and facial-attribute-preserving deformations. The modulated feature maps are rendered to a cartoon-style image with swapped fine layers. Our framework provides a multi-scale exaggeration control, and works with other StyleGAN-based image manipulation techniques. It creates realistic and detailed caricatures compared to other state-of-the-art methods.

\begin{figure}[t]
    \centering
    \normalsize
    \begin{tabularx}{\columnwidth}{@{\hspace{33pt}}c@{\hspace{0pt}}c}
    \includegraphics[width=0.37\columnwidth]{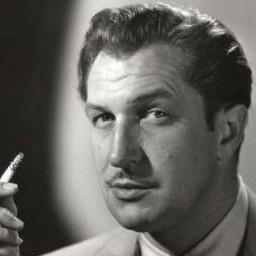}&
    \includegraphics[width=0.37\columnwidth]{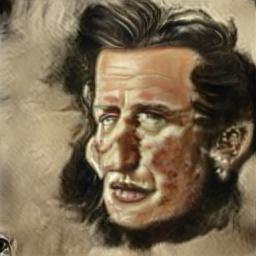} \\
    \vspace{-10.0pt}\\
    \includegraphics[width=0.37\columnwidth]{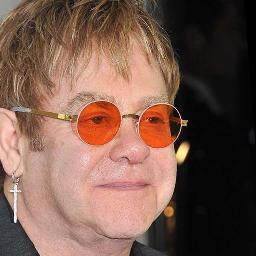} &
    \includegraphics[width=0.37\columnwidth]{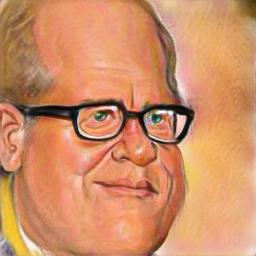} \\
    \vspace{-10.0pt}\\
    (a) & (b) \\
    \end{tabularx}
    \vspace{-0.3cm}
    \caption{\textit{Limitations.} Some types of input photos (a) lead to failed caricatures (b). The input on the top row is a gray-scale image, which results in unsuccessful photo-to-latent embedding because the training examples for the photo StyleGAN did not contain many gray-scale images. The failure in the embedding incurs an invalid output. The person on the bottom row wears orange sunglasses, but the output does not preserve this unique look. Besides, the hairstyle is changed as well. As we did not apply explicit supervision for glasses or hairstyles, such visual features may get removed in the generated caricatures. 
    {\footnotesize Inputs: \copyright Masheter Movie Archive/Alamy Stock Photo, \copyright Angela Weiss/Getty Images.}}
    \vspace{-0.1cm}
    \label{fig:limitation}
    \vspace{-0.5cm}
\end{figure}

\paragraph{Limitations.} In this paper, we did not explicitly handle data bias in the training dataset. Due to the bias, our system does not successfully preserve some traits of the input in certain cases (\Fig{limitation}). For example, the output caricature sometimes fails to preserve glasses contained in the input photo. In addition, some inputs that are largely different from the photo dataset, \emph{e.g.}, gray-scale images or photos under a dim light, may result in invalid shapes and cause visual artifacts. Handling the bias in the dataset in a systematic way is necessary for a more stabilized real-world automatic caricature system.
Besides, our caricature generator may disregard the hairstyle of the input subject under large deformations. Since our attribute matching loss only constrains the attributes related to facial shape, the preservation of hairstyle is not explicitly handled by supervision. This problem may be handled by adding new constraints on hairstyles, which is left as our future work.

\begin{figure*}
    \centering
    \large
    \begin{tabularx}{\textwidth}{@{\hspace{95pt}}c@{\hspace{0pt}}c@{\hspace{0pt}}c@{\hspace{0pt}}c}
    \includegraphics[align=c, width=0.17\textwidth]{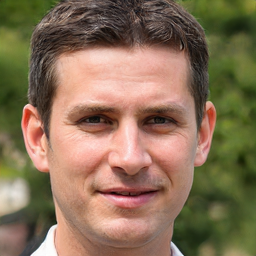} &
    \includegraphics[align=c, width=0.17\textwidth]{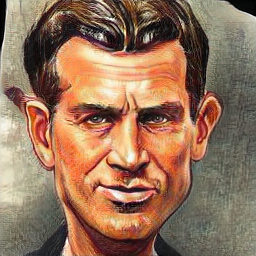} &
    \includegraphics[align=c, width=0.17\textwidth]{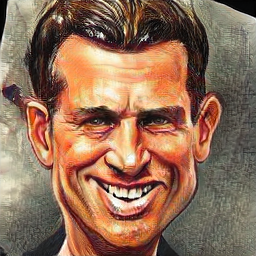} &
    \includegraphics[align=c, width=0.17\textwidth]{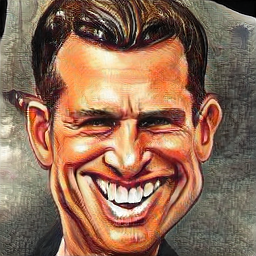} \\
    \includegraphics[align=c, width=0.17\textwidth]{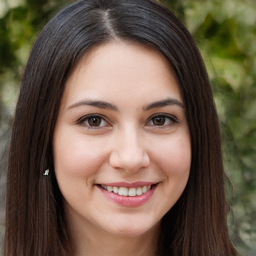} &
    \includegraphics[align=c, width=0.17\textwidth]{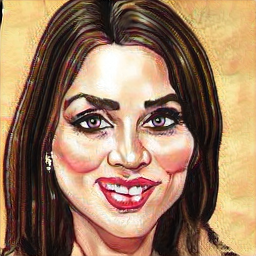} &
    \includegraphics[align=c, width=0.17\textwidth]{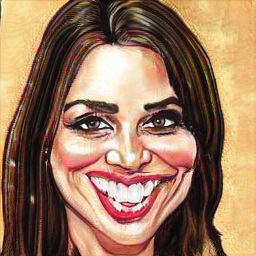} &
    \includegraphics[align=c, width=0.17\textwidth]{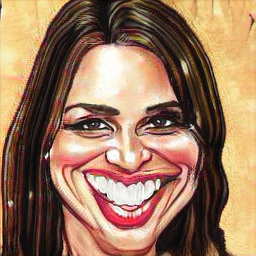} \\
    \includegraphics[align=c, width=0.17\textwidth]{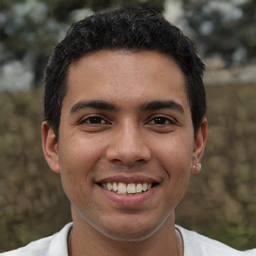} &
    \includegraphics[align=c, width=0.17\textwidth]{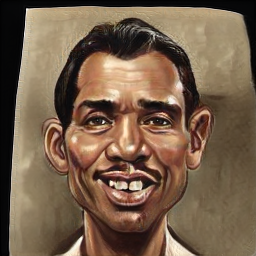} &
    \includegraphics[align=c, width=0.17\textwidth]{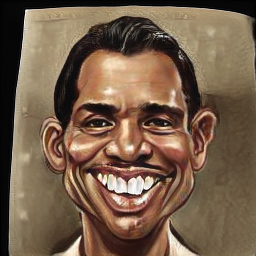} &
    \includegraphics[align=c, width=0.17\textwidth]{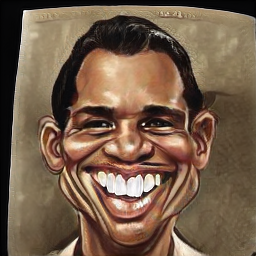} \\
    \vspace{-8.0pt}\\
    (a) & (b) & (c) & (d) \\
    \end{tabularx}
    \vspace{-0.3cm}
    \caption{\textit{Caricature expression manipulation.} Using a StyleGAN-based semantic image editing method, we can manipulate our caricature results. Starting from the input (a), caricatures are generated (b). We can add desired amount of smile increasingly (c, d). Note that the latent code editing direction to create smiles was searched using the plain StyleGAN trained for photos, but the edits create natural and realistic caricatures even with strong smiles.}
    \label{fig:app3}
    \vspace{-0.2cm}
\end{figure*}

